\documentclass{iopjournal}

\usepackage{amssymb}
\usepackage{latexsym}
\usepackage{amsmath,amsfonts}
\usepackage{array}
\usepackage{textcomp}
\usepackage{stfloats}
\usepackage{url}
\usepackage{verbatim}
\usepackage{mathtools}
\usepackage{float}
\usepackage{siunitx}
\usepackage[utf8]{inputenc}
\usepackage[english]{babel}
\usepackage{subcaption}
\usepackage{tablefootnote}
\usepackage{algorithm}
\usepackage{algpseudocode}
\newtheorem{lemma}{Lemma}
\newtheorem{theorem}{Theorem}

\begin{document}

\articletype{Article}

\title{Central Description Length (CDL) Clustering Validation Index}

\author{Mahdi Shamsi$^{1,*}$, Soosan Beheshti$^1$}

\affil{$^1$Toronto Metropolitan University, 350 Victoria St., Toronto, Ontario M5B 2K3, Canada}

\affil{$^*$Author to whom any correspondence should be addressed.}

\email{mahdi.shamsi@torontomu.ca}


\begin{abstract}
Selecting a clustering algorithm and its hyperparameters without labels is a common difficulty in engineering machine learning pipelines that work with unsupervised analysis of sensor, image, or process data. Clustering validation indices (CVIs) provide internal scores for ranking candidate clusterings, but most popular CVIs are built from Euclidean compactness and separation terms and so tend to favour compact, convex partitions. Their performance is known to degrade on non convex, irregular, or variable density data, where kernel transformations or alternative distance measures are typically used at the cost of additional tuning and computation. This paper introduces the Central Description Length (CDL) clustering validation index. CDL uses the observed within cluster compactness, the estimated cluster centers, and the estimated cluster covariances to compute a probabilistic upper bound on the description length associated with the unobservable true cluster centers. The bound condenses intra cluster compactness and centroid displacement into a single computable quantity and is evaluated on the partition produced by any clustering algorithm. The implementation uses only observable quantities (the data, the partition, the estimated centers, and the estimated covariances) and does not use ground truth labels. On synthetic benchmarks with non convex and arbitrary shape clusters, CDL-CVI selected the reference number of clusters more often and reached higher Adjusted Rand Index (ARI) values than the conventional CVIs we tested, without an additional kernel preprocessing stage. On image benchmarks (MNIST, CIFAR-10, STL-10) clustered from frozen unsupervised embeddings, CDL-CVI returned cluster numbers close to the reference class counts across K-means, DBSCAN, and spectral clustering in the reported trials. We also discuss limitations of the approach, in particular its dependence on covariance estimation, the chosen distance metric, and the input representation.
\end{abstract}

\section{Introduction}
Many engineering machine learning pipelines rely on the clustering of unlabeled measurements: fault diagnosis from vibration and acoustic signals, sensor state discovery in industrial processes, condition monitoring of mechanical and electrical systems, materials characterization, segmentation of images and signals, and exploratory grouping of process variables. In these settings labels are often expensive, partial, or unavailable, so the choice of clustering algorithm and its hyperparameters has to be made without ground truth \cite{jain2010, liu2010understanding}. Hyperparameters such as the number of clusters can change both the interpretability and the downstream utility of the clustering output \cite{bandyapadhyayfind, li2019clustering}. Clustering validation indices (CVIs) are internal scores used to rank candidate clusterings and to select hyperparameters by grid search \cite{hassan2024z, arbelaitz2013, gurrutxaga2011}.

The internal CVIs that are most often applied in practice are the Silhouette coefficient \cite{rousseeuw1987silhouettes}, the Davies and Bouldin (DB) index \cite{davies1979}, the Xie and Beni (XB) index \cite{xie1991validity}, the Dunn index \cite{dunn1973}, and the Calinski and Harabasz (CH) index \cite{calinski1974}. These indices combine an intra cluster compactness term with an inter cluster separation term, both of which are defined through Euclidean distances to cluster centroids or between pairs of points. Their structural form rewards small centroid to point distances and large centroid to centroid distances, and as a result they have known biases towards compact, well separated, near isotropic partitions \cite{liu2010understanding, arbelaitz2013}. They can therefore select inappropriate hyperparameters when the underlying clusters are non convex, have variable density, or lie on curved manifolds. Kernel transformations are a common workaround but introduce additional kernel and bandwidth tuning and increase the computational cost, especially in large datasets \cite{rahman2018}. Density Based Clustering Validation (DBCV) \cite{moulavi2014density} targets non convex, density based clusters via mutual reachability minimum spanning trees with an explicit treatment of noise, but is computationally heavy and sensitive to scaling in high dimensions. The Local Cores based Cluster Validity (LCCV) index \cite{cheng2018novel} uses high density local cores and graph distances and improves the behaviour with respect to cluster shape, but it relies on graph construction and shortest path computations and remains sensitive to outliers. Deep clustering \cite{khan2023multi, diallo2023auto, li2021contrastive} and multi view methods \cite{khan2024complementary} fold validation into joint representation learning, at the price of architectural choices and additional hyperparameters. The present paper is concerned with the more classical pipeline in which a clustering algorithm is applied to the data (or to a precomputed representation) and its hyperparameters are selected by an internal CVI.

We propose the Central Description Length Clustering Validation Index (CDL-CVI). The construction is motivated by the notion of $\epsilon$ Confidence Approximately Correct learning \cite{BeheshtiEcoac2025}, by the Minimum Noiseless Description Length (MNDL) principle \cite{Beheshti2005, fakhrzadeh2007minimum, grunwald2007minimum}, and by the Average Central Error \cite{beheshti2020k}. Description length and central error ideas in this family have been used previously for hyperparameter selection in linear regression, blind source separation, LTI system modeling, ARMA modeling, and subspace identification \cite{saba, tdelay, vedant, subid}. CDL-CVI uses the observed within cluster compactness together with the estimated cluster centers and covariances to compute a probabilistic upper bound on the description length associated with the unobservable true cluster centers. The resulting upper bound is the loss function: intra cluster compactness and the displacement between estimated and true centers are folded into a single quantity, instead of being combined as two separate terms. The implementation does not use ground truth labels; the true centers in the definition of the bound are theoretical latent quantities, and the operational index is computed from the data, the partition returned by the clustering algorithm, and the empirical cluster centers and covariances.

\paragraph{Relationship to prior CDL and MACE based methods.}
The central error description length idea underlying CDL-CVI was originally introduced for estimating the number of clusters in K-means type settings: the MACE means algorithm \cite{shahbaba2014}, K-MACE and kernel K-MACE \cite{beheshti2020k, rahman2018}, and the correct number of clusters description length criterion \cite{shamsi2019correct}. Those works tie the central error bound to a K-means objective and use it to choose $K$. The present paper repositions the same theoretical quantity as an internal CVI that is evaluated on the partition returned by any clustering algorithm (DBSCAN, OPTICS, spectral, hierarchical, K-means) and used to rank candidate hyperparameter settings, including settings in which the number of clusters is determined implicitly by other hyperparameters (for example DBSCAN's $\epsilon$ and $\min\textsc{Pts}$). It is not a new clustering algorithm. The bound itself is reused from \cite{beheshti2020k}; the new content of this paper is (i) recasting the bound as a CVI loss that is comparable across clustering families, (ii) defining the operational pipeline (covariance estimation, noise handling, validation and confidence probabilities, optimization direction), and (iii) an empirical comparison with widely used internal CVIs on synthetic non convex benchmarks and on image embeddings of the kind that arise in unsupervised feature pipelines in engineering applications.

\paragraph{Contributions.}
\begin{itemize}
    \item We formulate CDL-CVI, a single term loss function defined as a probabilistic upper bound on the description length of the latent cluster centers, obtained from the observed cluster compactness and the estimated within cluster covariances. The bound captures both intra cluster compactness and centroid displacement, without combining heterogeneous compactness and separation terms.
    \item We define an operational pipeline for CDL-CVI that applies to any clustering algorithm: per cluster center, compactness and covariance estimation, the treatment of noise points produced by density based methods, and a minimization objective with explicit validation and confidence probabilities. A pseudocode description is provided.
    \item We report experiments on non convex synthetic benchmarks (three rings, Path Based, Aggregation, Jain's Toy, D31, S15) for DBSCAN, OPTICS, and spectral clustering, and on image embeddings (MNIST, CIFAR-10, STL-10) for K-means, DBSCAN, and spectral clustering. CDL-CVI is compared with the Silhouette, DB, Dunn, XB, and CH indices on the same candidate hyperparameter grids, with a kernel based Silhouette baseline added on the synthetic experiments. The Adjusted Rand Index (ARI) and the Normalized Variation of Information (NVI) are reported as post selection external metrics.
\end{itemize}
The rest of the paper is organized as follows. Section~\ref{sec:valIdx} fixes notation for clustering, hyperparameters, and existing CVIs. Section~\ref{sec:CDLPro} develops CDL-CVI from the description length viewpoint and gives the operational pseudocode. Section~\ref{seq:sim} reports the synthetic and real data experiments. Section~\ref{sec:concl} discusses limitations and concludes.

\section{Validation Index in Clustering}
\label{sec:valIdx}
\subsection{Notations}
The unlabeled input dataset of length $n$ is represented as $\mathbf{x}_n= \left [ x(1), \dots, x(n) \right ]^T$. Each input data point, $x(i) \in \mathbf{x}_n$, is commonly referred to as a feature vector with dimension $d$, $x(i) \in \mathbb{R}^d$. The notations used in this paper are as follows:
\begin{itemize}
    \item $K^*$: True number of clusters
    \item $K$: Estimated Number of clusters by the clustering algorithm
    \item $c_{K}(k)$: $k^{\mathrm{th}}$ cluster in $K$ clustering 
    \item $\mathbf{C}_K = \{c_K (1), \dots, c_K(K) \}$: Set of clusters in $K$ clustering
    \item $\Theta = \{\theta_1, \theta_2, ..., \theta_m\}$: Hyperparameters of the clustering algorithm. The number of parameters, $m$, varies depending on the clustering algorithms.
\end{itemize} 

\subsection{Clustering Methods and Validation Indices}
The goal of a clustering algorithm is to partition the data into subsets, or clusters, such that the points within a cluster are more similar to each other than they are to points in other clusters. The notion of similarity is set by the algorithm itself, but its behaviour is also controlled by a small set of hyperparameters that must be fixed before clustering begins, and that strongly affect the resulting partition.
\begin{table}[H]
\centering
\caption{Clustering algorithms and their associated hyperparameters.}
\label{tab:cluster}
\begin{tabular}{|c|c|}
\hline
\begin{tabular}[c]{@{}c@{}}Clustering Algorithm\end{tabular} & Hyperparameters $\Theta$        \\ \hline
K-Means                                                        & $K$                         \\ \hline
DBSCAN, OPTICS                                                 & $\epsilon$, min samples     \\ \hline
Spectral Clustering                                            & $K$, kernel parameters      \\ \hline
Hierarchical Clustering                                        & affinity, Linkage           \\ \hline
BIRCH                                                          & threshold, branching factor \\ \hline
\end{tabular}
\end{table}
 Table \ref{tab:cluster} provides examples of such hyperparameters, $\Theta = \{\theta_1, \theta_2, \cdots, \theta_m\}$, for clustering methods such as K-means, DBSCAN, hierarchical and spectral algorithms. While $m$ is one for K-means, $m$ (the number of hyperparameters) is two for DBSCAN or OPTICS. As the table indicates, some clustering algorithms, such as K-means, require the number of clusters $K$ as an input to the algorithm and in such cases, the hyperparameter $\Theta$ includes the value of $K$. However, in several other clustering algorithms, the choice of hyperparameters indirectly determines the value of $K$. For instance, in DBSCAN, the selection of hyperparameters '$\epsilon$' and '\textit{min samples}' implicitly dictates the number of clusters that the algorithm produces. The clustering process provides the number of clusters $K$ as a function of hyperparameters ($K_\Theta$) and  the clustering result is denoted by $\mathbf{C}_{K_\Theta}$:
\begin{equation}
(\mathbf{x}_n, \Theta) \rightarrow K_\Theta, \mathbf{C}_{K_\Theta}(\mathbf{x}_n, \Theta)
\end{equation}
The resulted ${K_\Theta}$-clustering groups the data as follows:
\begin{align}
    \mathbf{C}_{K_\Theta} (\mathbf{x}_n, \Theta) \rightarrow \{ x_{{K_\Theta}k}(i) \in {c}_{K_\Theta}(k), k=1, \dots, K \} \label{eqn:cnkt}
\end{align}
In $x_{{K_\Theta}k}(i)$ the first subscript $K_\Theta$ is the total number of clusters and the second subscript $k$ refers to the $k$th cluster of the $K_\Theta$ clustering. Points that do not belong to any cluster are treated as outliers. Clustering validation indices (CVIs) are objective functions used to score clustering outputs. Each index summarizes the quality of a partition through some notion of intra cluster compactness and inter cluster separation: compactness measures how similar the points within a single cluster are, while separation measures how dissimilar points in different clusters are. CVIs are also used to compare different clustering algorithms on the same data, and most often to compare candidate hyperparameter settings of the same algorithm. The objective function of the CVI is written as
\begin{align}
   \mathrm{CVI\;objective\;function} = \ell_\mathrm{CVI}(\mathbf{C}_{K_\Theta}(\mathbf{x}_n, \Theta))
\end{align}
To find the optimum hyperparameter using this cost function, a set of possible hyperparameters are compared through a grid search:
\begin{align} \label{eq:5l}
    \Theta^*_{\ell_\mathrm{CVI}} = \underset{\Theta \in \boldsymbol{\Theta}}{\arg\min} \; \ell_{\mathrm{CVI}}(\mathbf{C}_{K_\Theta}(\mathbf{x}_n, \Theta))
\end{align}
where $\Theta^*_{\ell_\mathrm{CVI}}$ includes the optimal hyperparameters with respect to the loss function $\ell$ and $\boldsymbol{\Theta}$ is the set of all possible combinations of the competing hyperparameters. While the clustering algorithms and their corresponding hyperparameters are presented in Table \ref{tab:cluster}, well-established clustering validation indices objective functions are presented in Table \ref{tab:validation}.
\begin{table}[]
\centering
\caption{Clustering validation indices and their associated loss function.}
\label{tab:validation}
\resizebox{\textwidth}{!}{%
\begin{tabular}{|l|l|l|}
\hline
\begin{tabular}[c]{@{}l@{}}Validation\\ Index\end{tabular} &
  Cost Function $\ell(f(\mathbf{x}_n, \Theta))$ &
  Parameters Description \\ \hline
\begin{tabular}[c]{@{}l@{}}Silhouette\\ Coefficient\end{tabular} &
  $s_K = \frac{b - a}{\max\{a, b\}}$ &
  \begin{tabular}[c]{@{}l@{}}$a$: Mean distance between a \\ sample and all other points in the same class\\ $b$: Mean distance between a\\ sample and all other points\\ in the next nearest cluster\end{tabular} \\ \hline
\begin{tabular}[c]{@{}l@{}}Davies-Bouldin\\ Index\end{tabular} &
  \begin{tabular}[c]{@{}l@{}}$DB = \frac{1}{K} \sum_{i=1}^K \max_{i \neq j} R_{ij}$, $R_{ij}$\\ \\ $\;\;\;\;\; =  \frac{s_i + s_j}{d_{ij}}$\end{tabular} &
  \begin{tabular}[c]{@{}l@{}}$s_i$: Average distance between\\ each point of cluster and the\\ center of that cluster\\ $d_{ij}$: Distance between cluster\\ centers $i$ and $j$\end{tabular} \\ \hline
Dunn Index &
  $DI = \frac{\min_{1 \leq i < j \leq K} d(i, j)}{\max_{1 \leq k \leq K} \delta(k)}$ &
  \begin{tabular}[c]{@{}l@{}}$d(i,j)$:  Intercluster distance metric,\\ between clusters $i$ and $j$. \\ $\delta(k) =\underset{x(i), x(j) \in c_K(k)}{\max} {d(x(i), x(j))}$:\\ Variance between members of a cluster.\end{tabular} \\ \hline
\begin{tabular}[c]{@{}l@{}}Calinski-Harabasz\\ Index\end{tabular} &
  $CH(K) = \frac{\mathrm{tr}(B_K)}{\mathrm{tr}(W_K)} \times \frac{n - K}{K - 1}$ &
  \begin{tabular}[c]{@{}l@{}}$W_K$:within-cluster dispersion matrix\\ $B_K$:Between-group dispersion matrix\end{tabular} \\ \hline
\end{tabular}%
}
\end{table}

\section{Central Description Length Clustering Validation Index}
\label{sec:CDLPro}
Based on the provided notations in previous section, the following equations describe the clustering assumptions and notations:
\begin{align}
    &\mathbf{C}_{K^*}:  \forall x(i) \in \mathbf{x}_n: x(i) =\overline{ \mu}_{x(i)} + {\omega}_{x(i)}  \label{eqn:struc}
\end{align}
In (\ref{eqn:struc}) the true underlying clustering denoted by $\mathbf{C}_{K^*}$ has $K^*$ clusters. The cluster center 
$\overline{\mu}_{x(i)}$ is:
\begin{align}
    x(i) \in c_{K^*}(k):\;\overline{\mu}_{x(i)}=\overline{\mu}_{K^*}(k) = \frac{1}{n_{{K^*}k}} \sum^{n_{{K^*}k}}_{j=1}  x_{{K^*}k}(j) \label{eqn:muhat_2}
\end{align}
The available data point $x(i)$ in $\mathbf{x}_n$ is scattered by $\omega_{x(i)}$  which has a Gaussian distribution with variance $\overline{\sigma}^2_{x(i)} $\cite{beheshti2020k, shamsi2019correct}:
\begin{align} \label{eqn:trctr}
     x(i) \in c_{K^*}(k):\overline{\sigma}^2_{x(i)}=\overline{\sigma}^2_{K^*}(k) = {\mathrm{var}}(x_{{K^*}k}(i) - \overline{\mu}_{{K^*}}(k))
\end{align}

\begin{figure}[!htb]
\centering
\includegraphics[trim=0cm 0cm 0cm 0cm, scale=0.40]{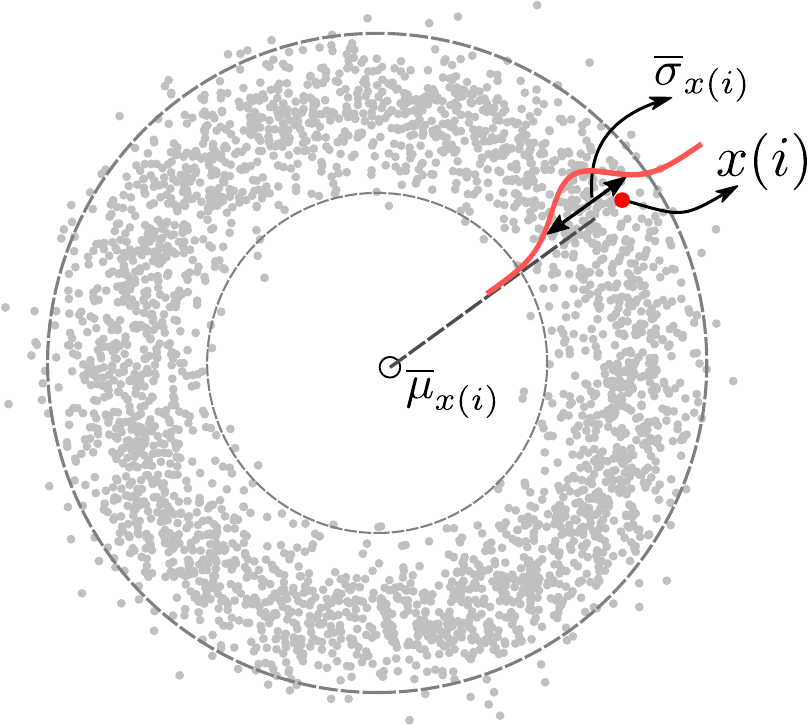}
\caption{An example of the $x(i)$ with the associated $\overline{\mu}_{x(i)}$ and $\overline{\sigma}_{x(i)}$}\label{fig:Fadcircles}
\label{fig:circle_GA}
\end{figure}

Figure~\ref{fig:Fadcircles} illustrates this structure. The $K_\Theta$ clustering in (\ref{eqn:cnkt}) returns one estimated center per cluster: the center of the $k$th cluster, denoted $\widehat{\mu}_{K_\Theta}(k)$, is the sample mean of the cluster's data points,
\begin{align}
    c_{K_\Theta}(k):\;\widehat{\mu}_{K_\Theta}(k) = \frac{1}{n_{{K_\Theta}k}} \sum^{n_{{K_\Theta}k}}_{i=1}  x_{{K_\Theta}k}(i) \label{eqn:muhat}
\end{align}
where $n_{{K_\Theta}k}$ denotes the size of the $k$-th cluster. The variance of a cluster of data points serves as a quantitative measure of the dispersion of the data around the mean or the center. The scatteredness factor of each cluster is calculated based on the variation of the data in each cluster \cite{shamsi2019correct}.
\begin{align}
     c_{K_\Theta}(k):\widehat{\sigma}^2_{{K_\Theta}}(k) = {\mathrm{var}}(x_{{K_\Theta}k}(i) - \widehat{\mu}_{{K_\Theta}}(k))
\end{align}
Consequently, for each data point and the clustering algorithm with the hyperparameters ${\Theta}$, we have:
\begin{align}
    &\mathbf{C}_{K_\Theta}: \forall x(i) \in c_{K_\Theta}(k), \widehat{\mu}_{x(i)} = \widehat{\mu}_{K_\Theta}(k), \widehat{\sigma}_{x(i)} = \widehat{\sigma}_{K_\Theta}(k) \label{eqn:estcs}\\
    &\mathbf{C}_{K_\Theta}: \forall x(i) \in \mathbf{x}_n, x(i) =\widehat{ \mu}_{x(i)} + {\omega'}_{x(i)} , {\omega'}_{x(i)} \sim \mathcal{N}(0, \widehat{\sigma}^2_{{x(i)}}) \label{eqn:struc22}
\end{align}
\begin{figure}[!htb]
\centering
\includegraphics[trim=0cm 0cm 0cm 0cm, scale=0.4]{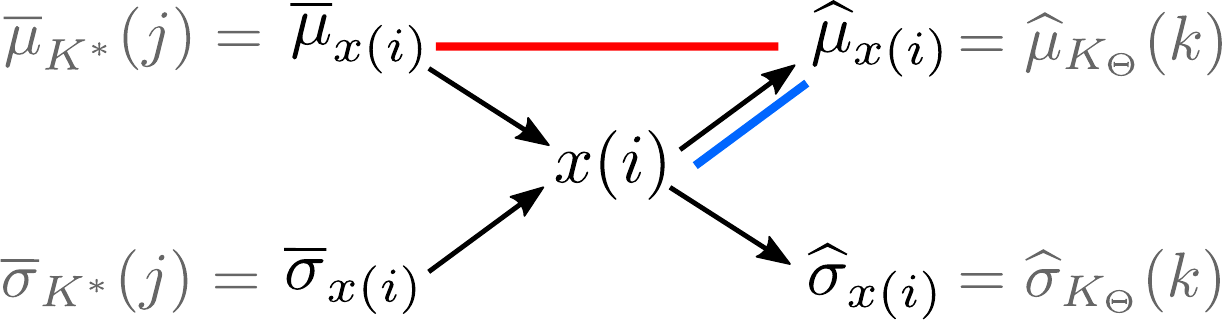}
\caption{Clustering procedure for data point $x(i) \in \mathbf{x}_n$, which belongs to the true cluster $c_{K^*}(j)$ and is assigned by the clustering algorithm to $c_{K_\Theta}(k)$. The solid blue line represents one distance contributing to the \emph{available} cluster compactness defined in equation~(\ref{eqn:CP}), and the solid red line represents one distance contributing to the \emph{unavailable} central error defined in equation~(\ref{eqn:CE}).}
\label{fig:va_digram}
\end{figure}
Figure~\ref{fig:va_digram} sketches the clustering procedure for a single data point. A point $x(i)$ that lies in the unknown true cluster $c_{K^*}(j)$, $1 \leq j \leq K^*$, is assigned by the algorithm to a cluster $c_{K_\Theta}(k)$, $1 \leq k \leq K_\Theta$, as in (\ref{eqn:estcs}). Given the estimated centers of the $K_\Theta$ clustering, the description length of the data is the average description length of the clustered data with respect to the estimated centers \cite{cover1999elements, rodrigues2021information},
\begin{align} \label{dl1}
{\mathop{\rm DL}} (\mathbf{x}_{{K_\Theta}};\widehat{\boldsymbol{\mu}}_{{K_\Theta}}) = \sum_{k=1}^{K_\Theta} \left ( \log_2 \sqrt{2\pi \widehat{\sigma}^2_{{K_\Theta}}(k)} \right ) + \gamma y_{{K_\Theta}}
\end{align}
where $\widehat{\boldsymbol{\mu}}_{{K_\Theta}}$ denotes the vector of cluster centers, $\gamma = \frac{\log_2 e}{2\widehat{\sigma}^2_{K_\Theta}}$ is a constant number \cite{fakhrzadeh2007minimum}, $\mathbf{x}_{{K_\Theta}}$ denotes the set of all the clustered data points and $y_{K_{\theta}}$ denotes the {\bf cluster compactness}:
\begin{equation}\label{eq:8}
y_{K_\Theta}  = \frac{1}{n} \sum_{k=1}^{K_\Theta}y_{{{K_\Theta}k}}
\end{equation}
where $y_{{K_\Theta}k}$ in (\ref{eq:8}) is the $k$-th cluster's compactness:
\begin{align}
    y_{{K_\Theta}k} = \sum_{i=1}^{n_{{K_\Theta}k}} \left \| x_{{K_\Theta}k}(i) - \widehat{\mu}_{{K_\Theta}}(k) \right \|_2^2 \label{eqn:CP}
\end{align}
Cluster compactness $y_{K_\Theta}$ is the most commonly used building block of clustering validation indices, and $y_{{K_\Theta}k}$ in (\ref{eqn:CP}) measures how tightly the points of cluster $k$ are packed around its center. A high value of $y_{{K_\Theta}k}$ corresponds to a loose cluster and a low value to a tight one. Cluster compactness is a monotonically decreasing function of the number of clusters: as the number of clusters grows, every point is closer on average to its own cluster center. The description length ${\mathop{\rm DL}} (\mathbf{x}_{{K_\Theta}};\widehat{\boldsymbol{\mu}}_{{K_\Theta}})$ in (\ref{dl1}) is built directly on $y_{K_\Theta}$ and inherits the same monotonicity. If one ranks hyperparameters by this quantity, the largest available number of clusters always wins. So neither cluster compactness nor the description length in (\ref{dl1}) can be used directly as a CVI. In the next subsection we instead use these quantities to estimate a different description length that is suitable as a CVI.

\subsection{Central Description Length}
As Figure \ref{fig:va_digram} shows, for each clustered data point there are four related quantities: the true center and the true scatter factor of (\ref{eqn:muhat_2}) and (\ref{eqn:trctr}), and the estimated center and the estimated scatter factor of (\ref{eqn:estcs}). The description length built on the cluster compactness (\ref{eqn:CP}) measures the distance shown by the blue line, which decreases with the number of clusters and ignores the underlying structure of the data. The distance shown by the red line, between the estimated center $\widehat{\mu}_{x(i)}$ and the true center $\overline{\mu}_{x(i)}$, behaves differently: it grows when the estimated number of clusters departs from the true number, in either direction.
\begin{figure*}[h]
\centering
\includegraphics[trim=0cm 0cm 0cm 0cm, scale=0.63]{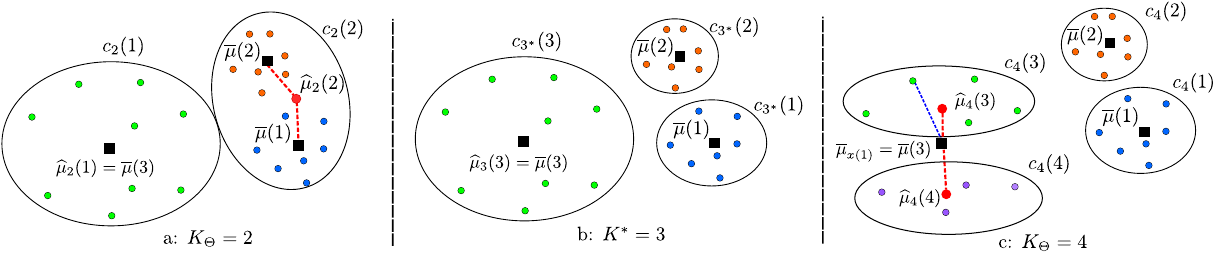}
\caption{Example of true centers and estimated centers for $K=2, 3, 4$ where $K^*=3$}
\label{fig:3sec}
\end{figure*}
Figure~\ref{fig:3sec} illustrates this for $K^*=3$ and three clustering scenarios with $K=2,3,4$. The average distance between $\widehat{\mu}$ and $\overline{\mu}$ is minimum at the true cluster number $K^*=3$ and grows when $K$ is smaller or larger than 3. A description length that reflects the variation between the true cluster centers and the estimated cluster centers should therefore be useful for selecting hyperparameters. We define the \textbf{Central Description Length (CDL)} as the average description length of the true cluster centers based on the estimated centers:
\begin{align}
    \label{eqn:CDL}
    {\mathop{\rm CDL}} ( \overline{\boldsymbol{\mu}}_{{{K_\Theta}}};\widehat{\boldsymbol{\mu}}_{{{K_\Theta}}}) = \sum_{k=1}^{K_\Theta} \left ( \log_2 \sqrt{2\pi \widehat{\sigma}_{{K_\Theta}}^2(k)}\right ) + \gamma' z_{{K_\Theta}}
\end{align} 
where $\overline{\boldsymbol{\mu}}_{{K_\Theta}} = \left [ \overline{\mu}_{K_\Theta}(1), \hdots, \overline{\mu}_{K_\Theta}(k), \hdots, \overline{\mu}_{K_\Theta}(K_\Theta) \right ]$ is the one-to-one corresponding vector of the (latent) true cluster centers of the clustered data points (has the same size as the data size itself), $\gamma' = \frac{\log_2 e}{2\overline{\sigma}^2}$ is a constant determined by the average true scatter, and $ z_{{K_\Theta}}$ is the \textbf{Average Central Error (ACE)} \cite{beheshti2020k}, defined as the average squared distance between the estimated cluster centers and the unavailable true cluster centers:
\begin{equation} \label{eq:6}
z_{K_\Theta}  = \frac{1}{n} \sum_{k=1}^{K_\Theta}z_{{{K_\Theta}k}}
\end{equation}
where $z_{{{K_\Theta}k}}$ in (\ref{eq:6}) denotes the \textbf{central error} for each cluster and is defined as the distance between the true associated center of each data point in the cluster and the estimated center of that cluster.
\begin{equation}
    z_{{K_\Theta}k} = \sum_{i=1}^{n_{{K_\Theta}k}} \left \| \overline{\mu}_{x_{{K_\Theta}k}(i)} - \widehat{\mu}_{{K_\Theta}}(k) \right \|_2^2 \label{eqn:CE}
\end{equation}

A direct calculation of ACE is not possible because $\overline{\mu}_{x(i)}$ is unknown, but it is shown in \cite{rahman2018, shahbaba2014} that the observable cluster compactness $y_{{K_\Theta}}$ can be used to construct probabilistic bounds on ACE. The connection between the cluster compactness and ACE distributions is established in \cite{beheshti2020k, shamsi2019correct}:
\begin{align}
    \underline{z_{K_\Theta}}(y_{K_\Theta}) \leq z_{{K_\Theta}} \leq \overline{z_{{K_\Theta}}}(y_{K_\Theta})
\end{align}
The derivation is summarized in the appendix and given in full in \cite{beheshti2020k, shamsi2019correct}. The Gaussian scatter assumption is used to obtain a closed form for the variance of the cluster compactness. When the scatter is non Gaussian the same expressions hold asymptotically by central limit arguments, provided the clusters are large enough and the scatter factors have finite moments. The bound is less tight in small clusters, with heavy tailed data, and in high dimensional embeddings where covariance estimation is unstable.

\subsection{Central Description Length Clustering Validation Index (CDL-CVI)}
\label{sec:cdl-cvi}
Bounds on CDL derived from the bounds on ACE in (\ref{eq:6}) were originally proposed for estimating the number of clusters $K$ in K-means type clustering \cite{beheshti2020k, shamsi2019correct, shahbaba2014}. In the present work CDL is evaluated on the partition produced by any clustering algorithm (DBSCAN, OPTICS, spectral, hierarchical, K-means, and so on) and used as an internal validation index for ranking hyperparameter settings, and where appropriate for comparing different clustering methods on the same data.

The true cluster centers $\overline{\mu}_{x(i)}$ and the latent partition $\mathbf{C}_{K^*}$ appear in the definition of the target (\ref{eqn:CDL}), but only as theoretical objects. CDL-CVI does not access any ground truth labels at run time. The computable upper bound $\overline{z_{K_\Theta}}$ depends only on the data $\mathbf{x}_n$, the partition $\mathbf{C}_{K_\Theta}$ returned by the clustering algorithm, the estimated cluster centers $\widehat{\mu}_{K_\Theta}(k)$, the observed within cluster compactness $y_{K_\Theta}$, and the estimated within cluster covariances $\widehat{\Sigma}_{K_\Theta}(k)$. External labels, when available, are used only after hyperparameter selection for the post hoc ARI and NVI evaluation.

Using the estimated upper bound on CDL as the validation index, the cost function in (\ref{eq:5l}) becomes
\begin{align}
    \ell_{\mathrm{CDL\text{-}CVI}}(\mathbf{C}_{K_\Theta}(\mathbf{x}_n, \Theta)) = \overline{z_{{K_\Theta}}} \label{eqn:zkthetals}
\end{align}
and the selected hyperparameter set is
\begin{align}
\label{opt_m}
\Theta^*_{\mathrm{CDL\text{-}CVI}} = \underset{\Theta \in \boldsymbol{\Theta}}{\arg\min}\;\overline{z_{{K_\Theta}}}.
\end{align}
The upper bound $\overline{z_{K_\Theta}}$ is the sum over clusters of the per cluster upper bounds derived in \cite{beheshti2020k}; the derivation is summarized in the appendix. Its computation involves two Chebyshev steps. The observed cluster compactness $y_{K_\Theta k}$ is first used to upper bound the unknown variation of the true centers within the estimated $k$th cluster, $\|\Delta_{K_\Theta k}\|_F^2$, with validation probability $P_v=1-1/\alpha_k^2$. The resulting upper bound is then propagated through the closed form for the mean and variance of the central error to obtain
\begin{align}
\label{eq:upperbound}
\overline{z_{K_\Theta}} \;=\; \sum_{k=1}^{K_\Theta} \overline{E[Z_{K_\Theta k}]} \;+\; \beta\,\sqrt{\mathrm{Var}[Z_{K_\Theta}]}\,,
\end{align}
with confidence probability $P_c=1-1/\beta^2$. The terms $\overline{E[Z_{K_\Theta k}]}$ and $\mathrm{Var}[Z_{K_\Theta}]$ are evaluated as in Lemmas 1 and 2 of the appendix, using the cluster size $n_{K_\Theta k}$ and the estimated within cluster covariance $\widehat{\Sigma}_{K_\Theta}(k)$ in place of the unknown $\overline{\Lambda}$. In the reported experiments we use the three sigma values $P_v = P_c = 0.997$ of \cite{beheshti2020k}.

The closed form expressions for $E[Y_{K_\Theta k}]$ and $E[Z_{K_\Theta k}]$ require zero mean scatter factors with finite second moments, and within cluster independence. The variance expression in Lemma~2 is derived under a Gaussian scatter assumption and holds asymptotically under non Gaussian scatter via central limit arguments, provided the clusters are large enough and the scatter factors have finite moments. The bound is conservative for small clusters, for heavy tailed data, and for high dimensional embeddings with unstable covariance estimates, as we discuss in the limitations. All squared distances in this paper are Euclidean, so the results depend on the chosen representation and on feature scaling; the bound should be applied after standardization or after a problem appropriate metric (for example cosine on normalized embeddings) has been chosen.

In practice, for each candidate hyperparameter $\Theta$ in the search grid $\boldsymbol{\Theta}$, the clustering algorithm is run on $\mathbf{x}_n$ and the resulting partition $\mathbf{C}_{K_\Theta}$ is read off. Empty clusters are dropped. Singleton clusters (size $1$) have zero compactness and a degenerate covariance; in this case the cluster covariance is replaced by a small ridge regularized estimate based on the dataset's pooled covariance, and the cluster contributes only its mean term to the bound. For density based clustering algorithms (DBSCAN, OPTICS) that label points as noise, the noise points are treated as unassigned and excluded from the internal CDL-CVI center, compactness, and covariance estimation. The noise fraction is reported alongside the score, and the same noise convention is applied to the baseline CVIs. Tables that involve DBSCAN or OPTICS state whether noise is excluded from the post selection ARI and NVI computation. Candidate hyperparameters are ranked by $\overline{z_{K_\Theta}}$, with smaller values preferred. Algorithm~\ref{alg:cdlcvi} summarizes the procedure.

\begin{algorithm}[H]
\caption{CDL-CVI hyperparameter selection}
\label{alg:cdlcvi}
\begin{algorithmic}[1]
\Require Data $\mathbf{x}_n$, clustering algorithm $\mathcal{A}$, hyperparameter grid $\boldsymbol{\Theta}$, validation probability $P_v$, confidence probability $P_c$
\Ensure Selected hyperparameter $\Theta^*$
\State $\alpha \gets 1/\sqrt{1-P_v}$, $\beta \gets 1/\sqrt{1-P_c}$
\ForAll{$\Theta \in \boldsymbol{\Theta}$}
    \State $\mathbf{C}_{K_\Theta} \gets \mathcal{A}(\mathbf{x}_n, \Theta)$ \Comment{run clustering}
    \State Drop empty clusters; exclude points labelled as noise
    \ForAll{cluster $k = 1, \dots, K_\Theta$}
        \State Compute $\widehat{\mu}_{K_\Theta}(k)$, compactness $y_{K_\Theta k}$ (\ref{eqn:CP}), and covariance $\widehat{\Sigma}_{K_\Theta}(k)$
        \State If singleton, regularize $\widehat{\Sigma}_{K_\Theta}(k)$ with ridge from pooled covariance
        \State Upper-bound $\|\Delta_{K_\Theta k}\|_F^2$ from $y_{K_\Theta k}$ using Chebyshev with $\alpha$
        \State Compute $\overline{E[Z_{K_\Theta k}]}$ via Lemma 1 using $\widehat{\Sigma}_{K_\Theta}(k)$
    \EndFor
    \State $\overline{z_{K_\Theta}} \gets \sum_k \overline{E[Z_{K_\Theta k}]} + \beta\sqrt{\mathrm{Var}[Z_{K_\Theta}]}$ (eq.~\ref{eq:upperbound})
\EndFor
\State \Return $\Theta^* \gets \arg\min_{\Theta \in \boldsymbol{\Theta}} \overline{z_{K_\Theta}}$
\end{algorithmic}
\end{algorithm}

For each clustering algorithm and dataset, every CVI in the comparison is evaluated on the same candidate clusterings produced by the same hyperparameter grid. Indices that are conventionally maximized (Silhouette, Dunn, Calinski and Harabasz) are converted to minimization losses by sign flip. External labels, when available, are used only after hyperparameter selection to compute ARI and NVI; the selection itself uses only the internal CVI.
\section{Simulation Results}
\label{seq:sim}
For every algorithm and dataset, the same candidate hyperparameter grid is used to generate the clustering outputs that are then scored by all CVIs. Selection is performed using only the internal CVI; the external scores ARI and NVI are computed after selection from the ground truth labels and play no role in the selection itself. ARI is reported on the standard scale where higher is better (perfect agreement gives ARI~$=1$). NVI is the normalized variation of information and lower is better (NVI~$=0$ indicates full agreement). For density based methods (DBSCAN, OPTICS), points labelled as noise are excluded from the internal CVI computation and from the ARI and NVI computations, and the noise fraction is reported separately when relevant. The same noise convention is applied to all baseline CVIs. In the synthetic tables the reference cluster count is the dataset's ground truth class count; in the real data tables the cluster numbers in parentheses refer to the reference class counts of the corresponding benchmark labels. The cluster number in each row is bolded when it matches the reference count, and the best ARI and best (lowest) NVI in each row are bolded independently. We set the validation and confidence probabilities to $P_v = P_c = 0.997$, the three sigma values of \cite{beheshti2020k}.

\subsection{Synthetic Dataset}
We evaluate CDL-CVI on two synthetic scenarios: a challenging three rings dataset \cite{motallebi2022local} and a collection of arbitrary shape datasets \cite{lee2018}. All experiments in this subsection were run on an AMD Ryzen 7 4800HS CPU with 8 cores and 16~GB of RAM. In the three rings dataset the three clusters are non convex and not linearly separable, and the two inner rings are close to one another, which makes the problem hard for both clustering algorithms and CVIs. The arbitrary shape datasets have irregular or intertwined cluster structures that violate the isotropic cluster assumption underlying many classical algorithms; many widely used CVIs are biased towards compact, well separated geometries and select poor hyperparameters on these datasets.

For non convex and arbitrary shape clusters, DBSCAN, OPTICS, and spectral clustering are the algorithms most often used in practice. K-means and Gaussian mixture models implicitly assume convex clusters and can only be applied to such data through a kernel transformation, which adds a kernel choice and bandwidth tuning step and is computationally expensive on large or high dimensional data.

\begin{figure}[!htb]
\centering
\includegraphics[trim=0cm 0cm 0cm 0cm, scale=0.37]{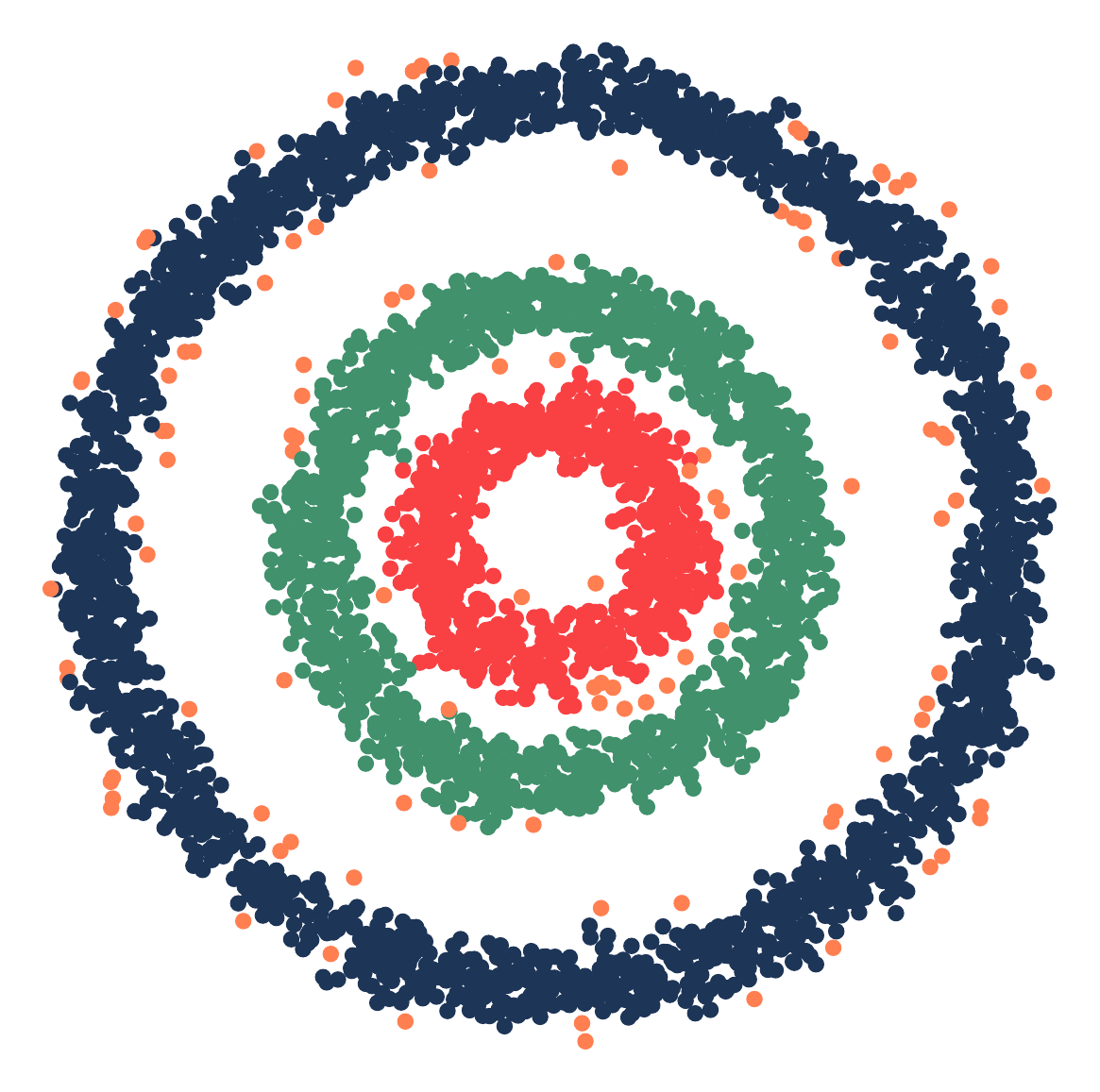}
\caption{CDL-CVI for selection of hyperparameter $\epsilon$ in DBSCAN}
\label{fig:DBSCAN_RE}
\end{figure}

\begin{figure}[htbp]
  \setlength{\belowcaptionskip}{-1pt} 
  \setlength{\abovecaptionskip}{5pt}  
  \setlength{\intextsep}{2pt}         
  \setlength{\floatsep}{-10pt}          
  \setlength{\textfloatsep}{5pt}      
    \centering
    \begin{subfigure}{0.4\columnwidth}
        \includegraphics[width=1\columnwidth, trim={0cm 0cm 0cm 0cm}, clip]{DBSCAN_1-eps-converted-to.pdf}
        \caption{$\epsilon=0.57$ (Proposed CDL-CVI)}
        \label{fig:db045}
    \end{subfigure}
    \begin{subfigure}{0.4\columnwidth}
        \includegraphics[width=1\columnwidth, trim={0cm 0cm 0cm 0cm}, clip]{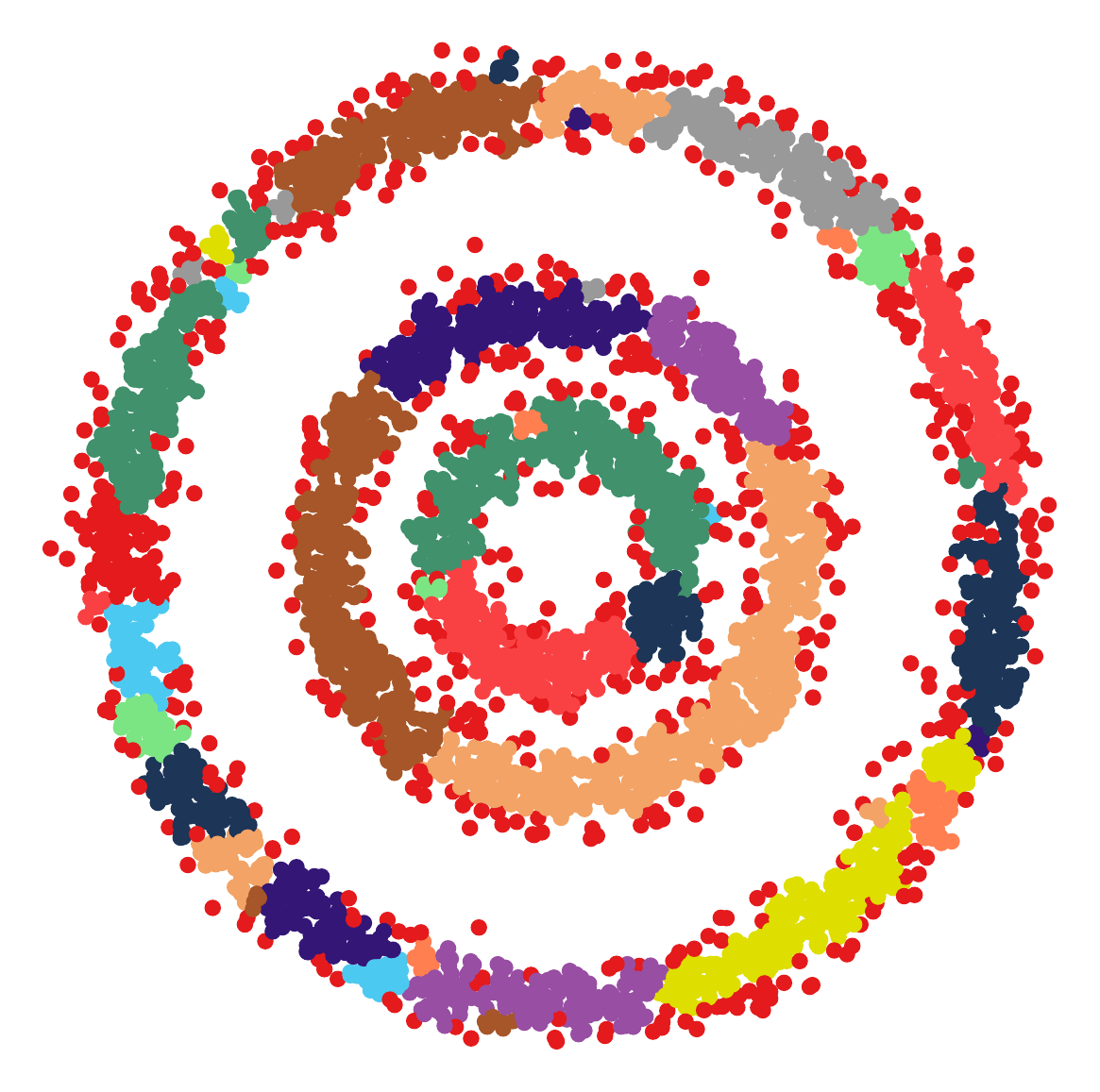}
        \caption{$\epsilon=0.45$ (Silhouette index; Dunn index selected a similar value)}
        \label{fig:db057}
    \end{subfigure}
    \vspace{2mm}
    \begin{subfigure}{0.4\columnwidth}
        \includegraphics[width=1\columnwidth, trim={0cm 0cm 0cm 0cm}, clip]{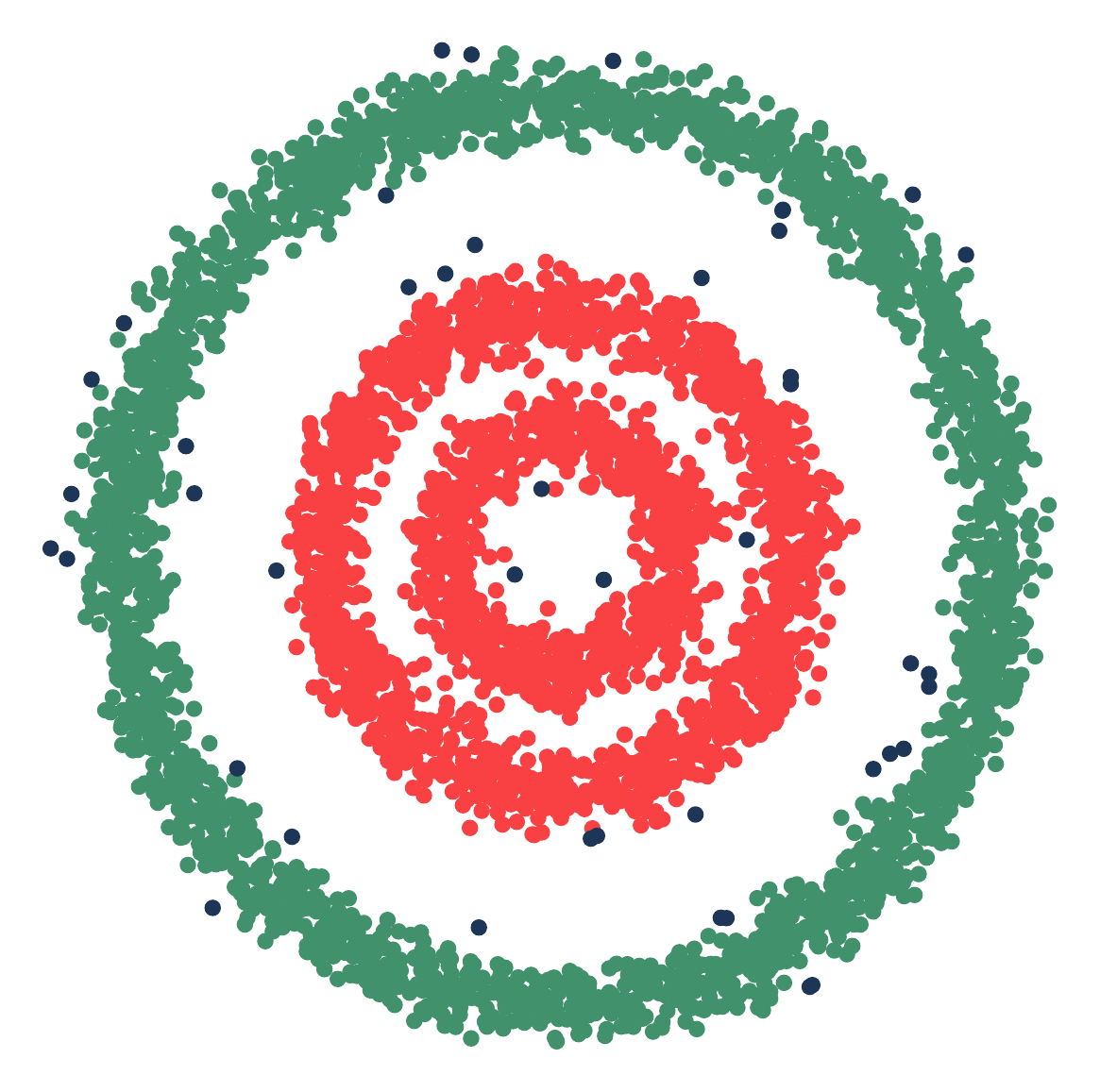}
        \caption{$\epsilon=0.87$ (Calinski-Harabasz and Davies-Bouldin index)}
        \label{fig:db075}
    \end{subfigure}
    \caption{CVI $\epsilon$ parameters tuning results in DBSCAN clustering}
    \label{fig:dbcompare}
\end{figure}

The inner ring of the three rings dataset is generated with radius 4, standard deviation 0.7, and 800 points; the middle ring with radius 8, standard deviation 0.7, and 1500 points; and the outer ring with radius 15, standard deviation 0.7, and 2500 points. We use DBSCAN for this experiment because it handles complex shapes well and tolerates noise. DBSCAN has two hyperparameters: the minimum number of points minPts, which is not very sensitive and is usually set to 4 or 5 (we use 5), and $\epsilon$, which controls the scale of clustering and is highly sensitive to the density and distribution of the data \cite{khan2014dbscan}. Figure~\ref{fig:DBSCAN_RE} shows $\overline{z_{K_\Theta}}$ from (\ref{eqn:zkthetals}) as a function of $\epsilon$ when CDL-CVI is used as in (\ref{opt_m}) to select $\epsilon$ over the range $\epsilon \in [0.45, 0.89]$, with the three sigma values $P_v = P_c = 0.997$ of \cite{beheshti2020k}. The bound is essentially flat for $0.54 < \epsilon < 0.72$ and attains its minimum at $\epsilon = 0.57$.
Figure~\ref{fig:db045} shows the DBSCAN partition at the $\epsilon$ chosen by CDL-CVI. The three rings are correctly recovered and a small number of points are flagged as outliers. The Silhouette and Dunn indices choose $\epsilon$ values of $0.45$ and $0.47$ respectively. Figure~\ref{fig:db057} shows the DBSCAN partition at $\epsilon = 0.45$ (the result for $\epsilon = 0.47$ is almost identical): DBSCAN now breaks each ring into many small dense subclusters. Figure~\ref{fig:db075} shows the DBSCAN partition at $\epsilon = 0.87$, the value chosen by both the Calinski-Harabasz and Davies-Bouldin indices: in this case the two inner rings are merged into a single cluster, so only two clusters are recovered instead of three.

\begin{figure}[htbp]
  \centering
  \begin{subfigure}[b]{0.45\textwidth}
    \centering
    \includegraphics[width=\textwidth]{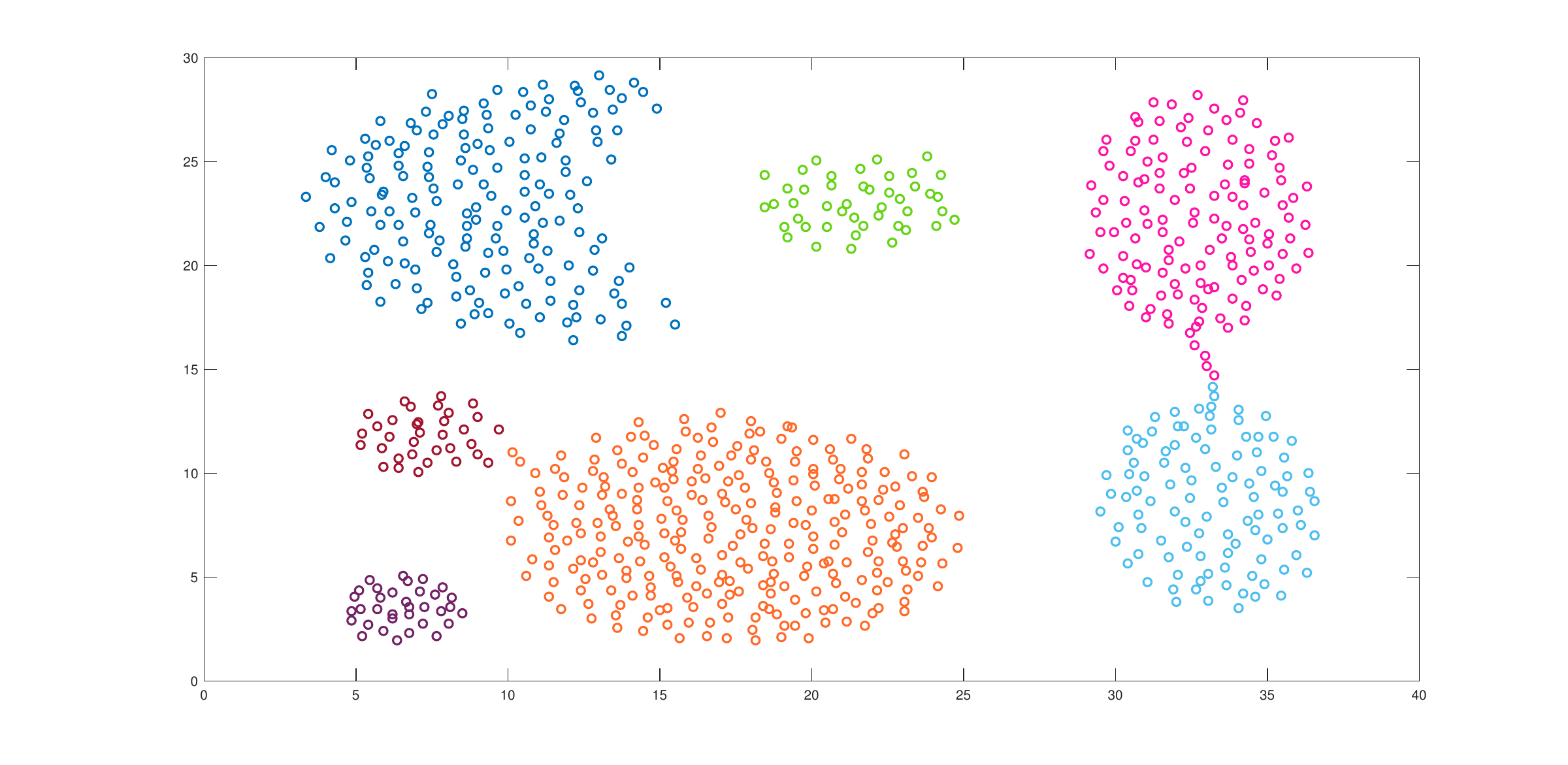}
    \caption{}
    \label{a:1}
  \end{subfigure}
  \hspace{-5.4mm}
  \begin{subfigure}[b]{0.45\textwidth}
    \centering
    \includegraphics[width=\textwidth]{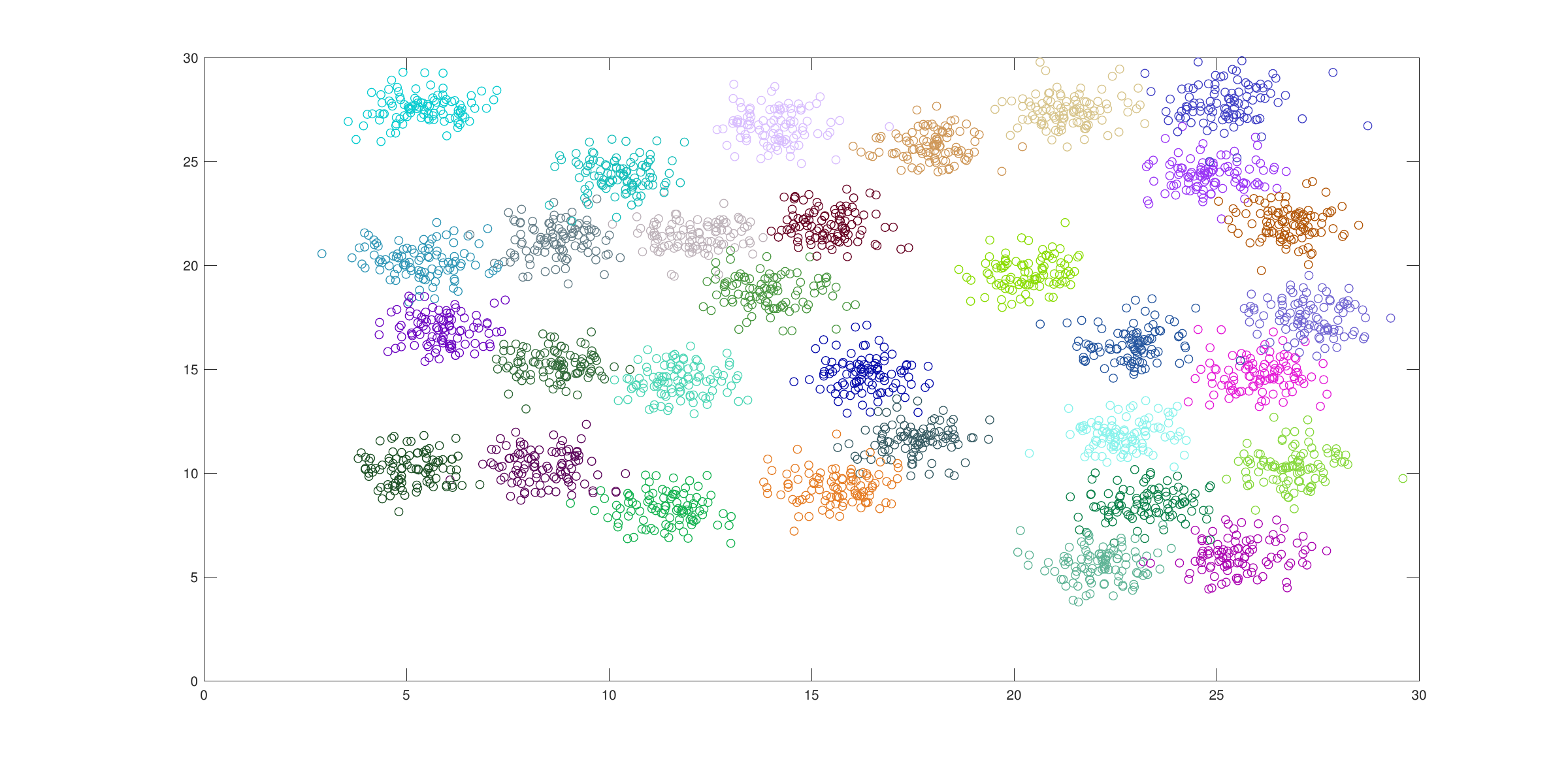} 
    \caption{}
    \label{b:1}
  \end{subfigure}
  \hspace{-5.4mm}
  \begin{subfigure}[b]{0.45\textwidth}
    \centering
    \includegraphics[width=\textwidth]{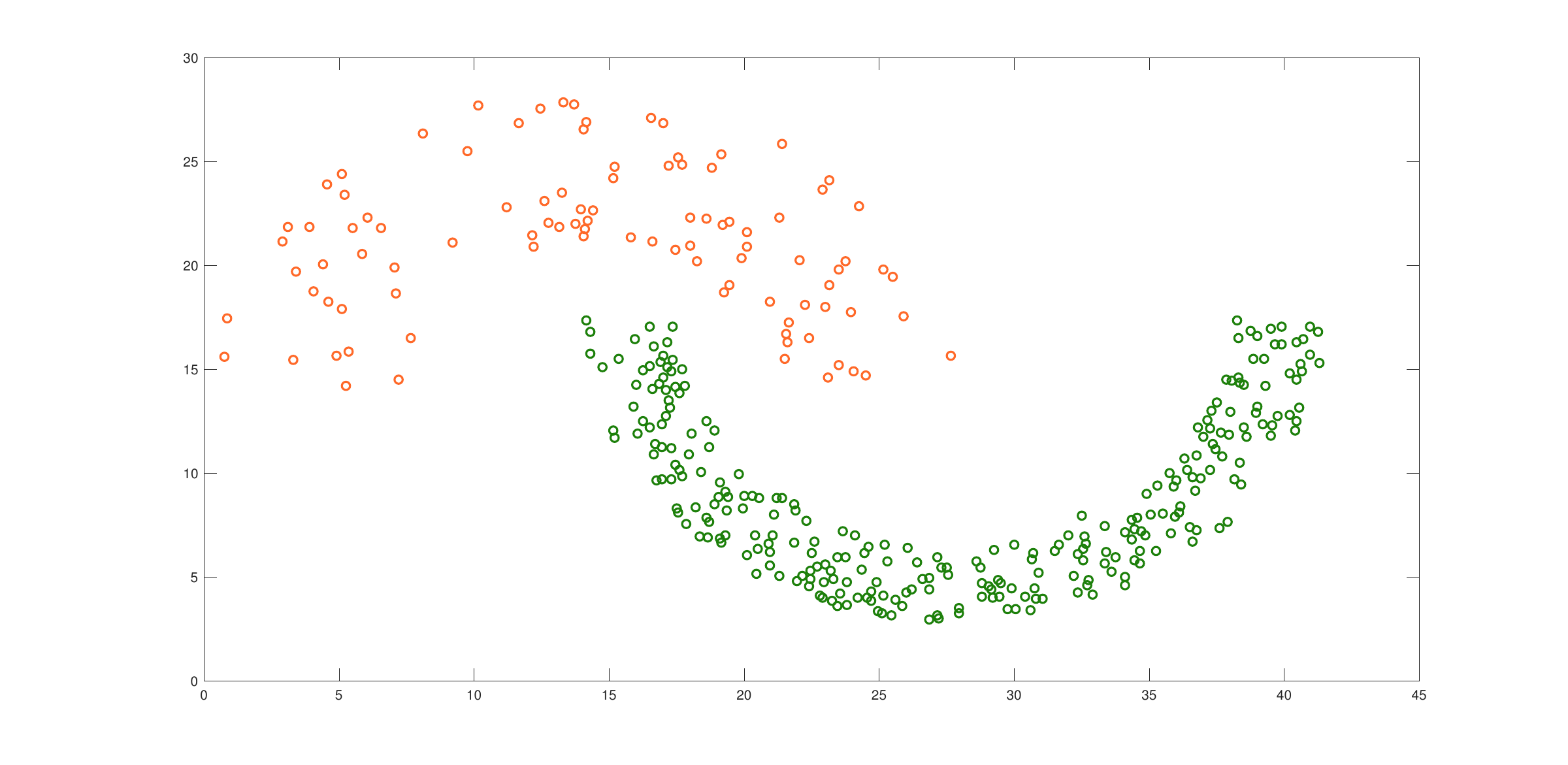} 
    \caption{}
    \label{c:1}
  \end{subfigure}
  \hspace{-5.4mm}
  \begin{subfigure}[b]{0.45\textwidth}
    \centering
    \includegraphics[width=\textwidth]{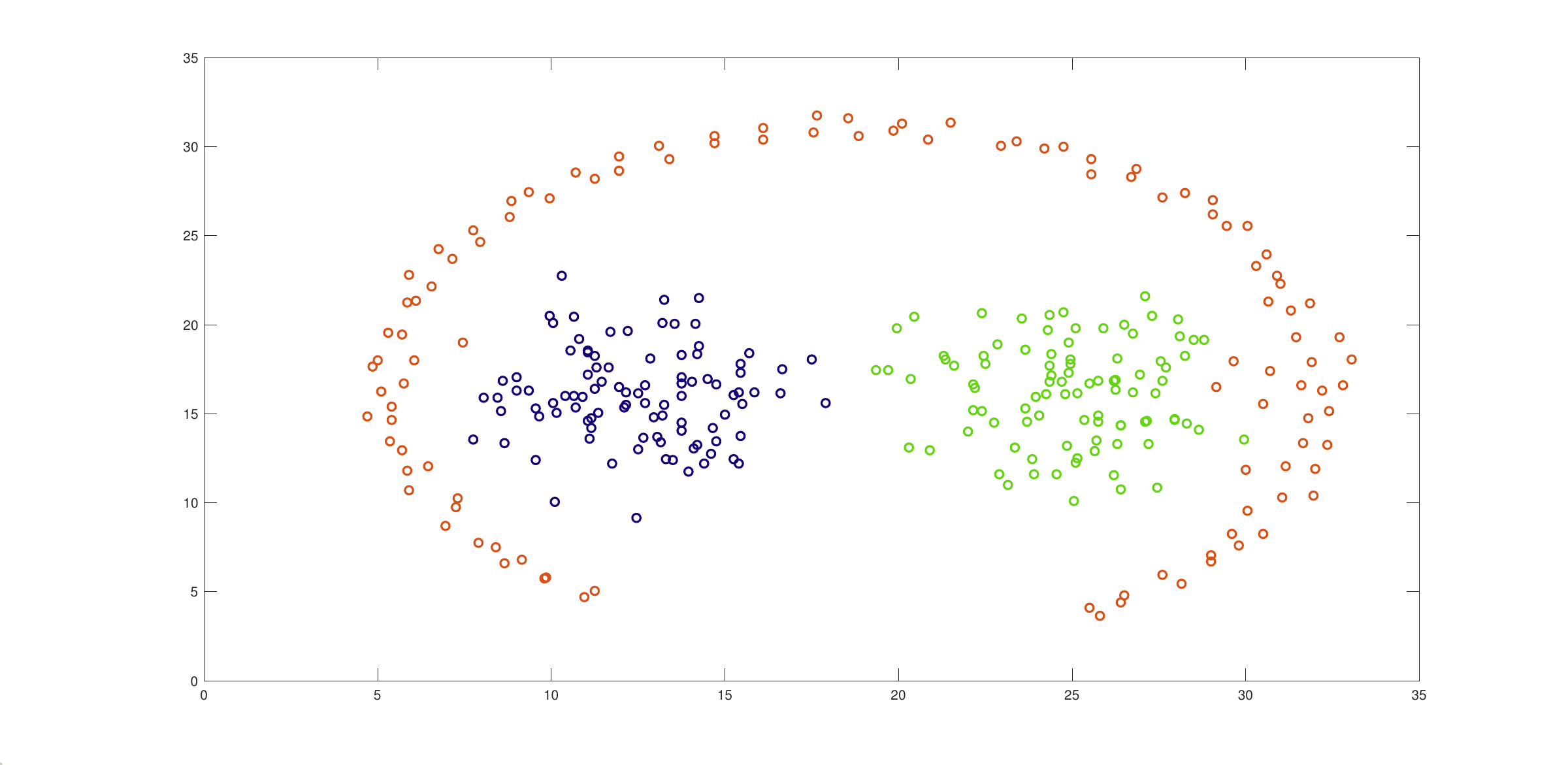} 
    \caption{}
    \label{d:1}
  \end{subfigure}
  \hspace{-5.4mm}
  \begin{subfigure}[b]{0.45\textwidth}
    \centering
    \includegraphics[width=\textwidth]{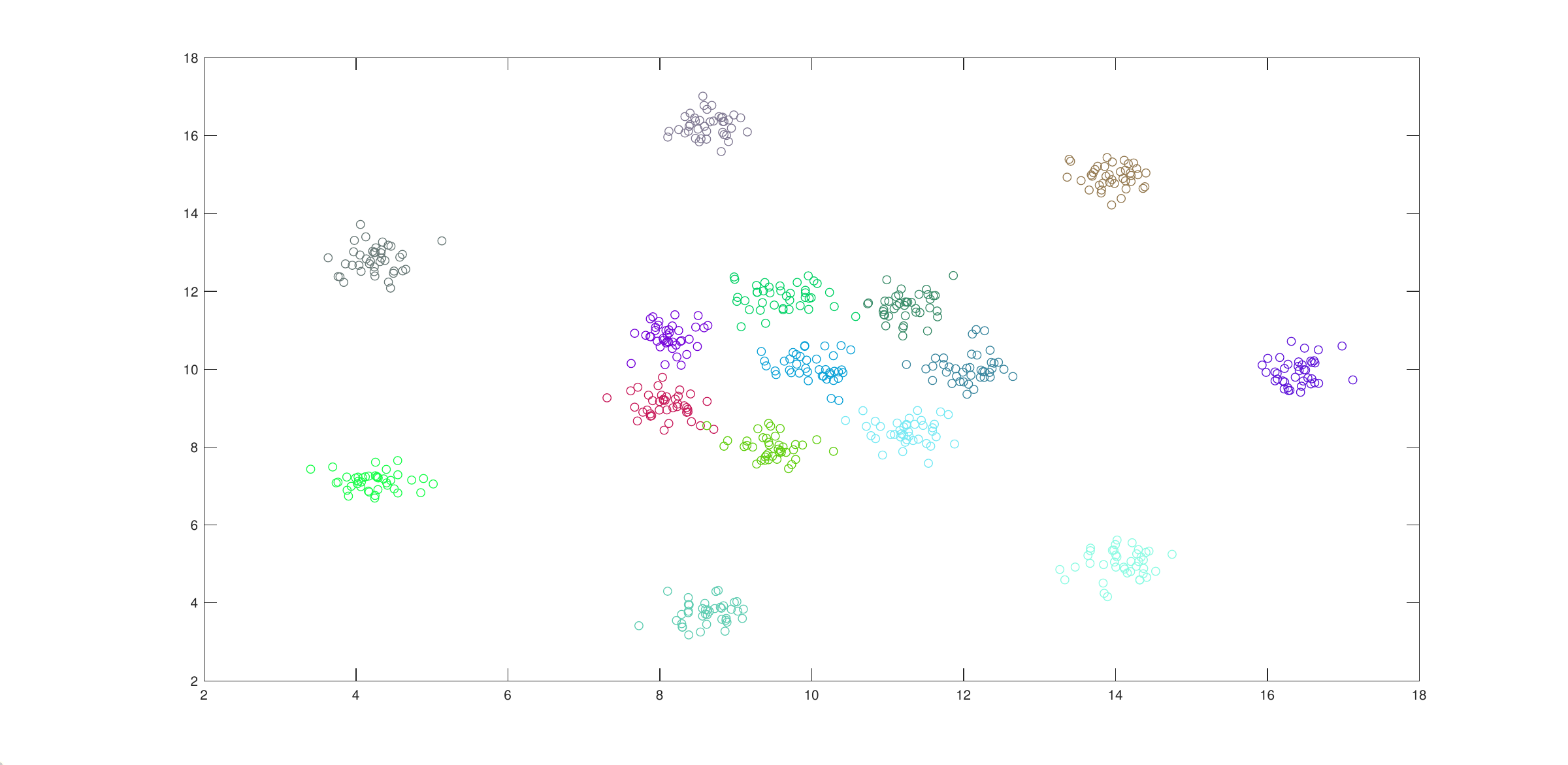} 
    \caption{}
    \label{e:1}
  \end{subfigure}
  \hspace{-5.4mm}
  \caption{Benchmark datasets. (a) Aggregation, (b) D31, (c) Jain's Toy, (d) Path-Based, (e) S15.} \label{fig:2-sim}
\end{figure}

\begin{figure}[htbp]
  \centering
  \begin{subfigure}[b]{0.45\textwidth}
    \centering
    \includegraphics[width=\textwidth]{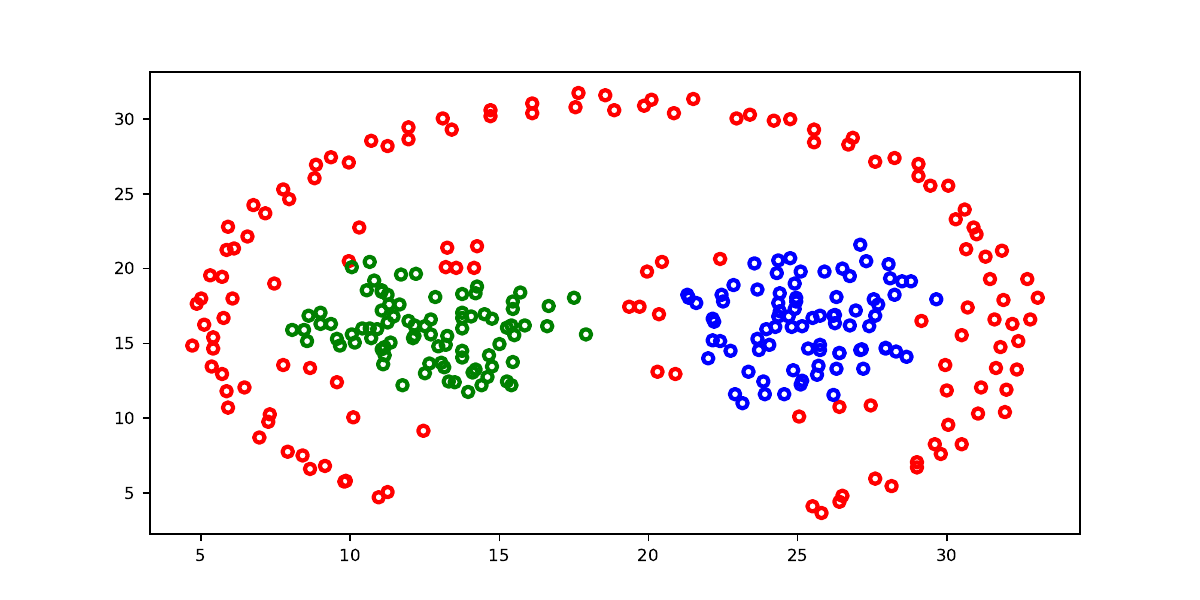} 
    \caption{Proposed CDL index}
    \label{a}
  \end{subfigure}
  \hspace{-5.4mm}
  \begin{subfigure}[b]{0.45\textwidth}
    \centering
    \includegraphics[width=\textwidth]{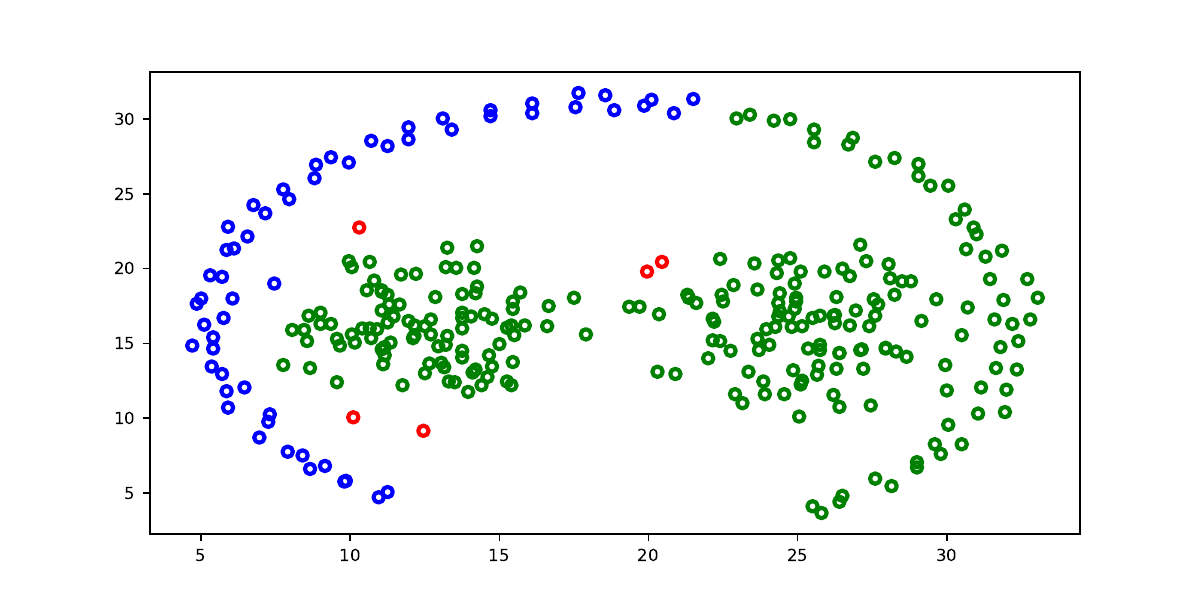} 
    \caption{DB index}
    \label{b}
  \end{subfigure}
  \hspace{-5.4mm}
  \begin{subfigure}[b]{0.45\textwidth}
    \centering
    \includegraphics[width=\textwidth]{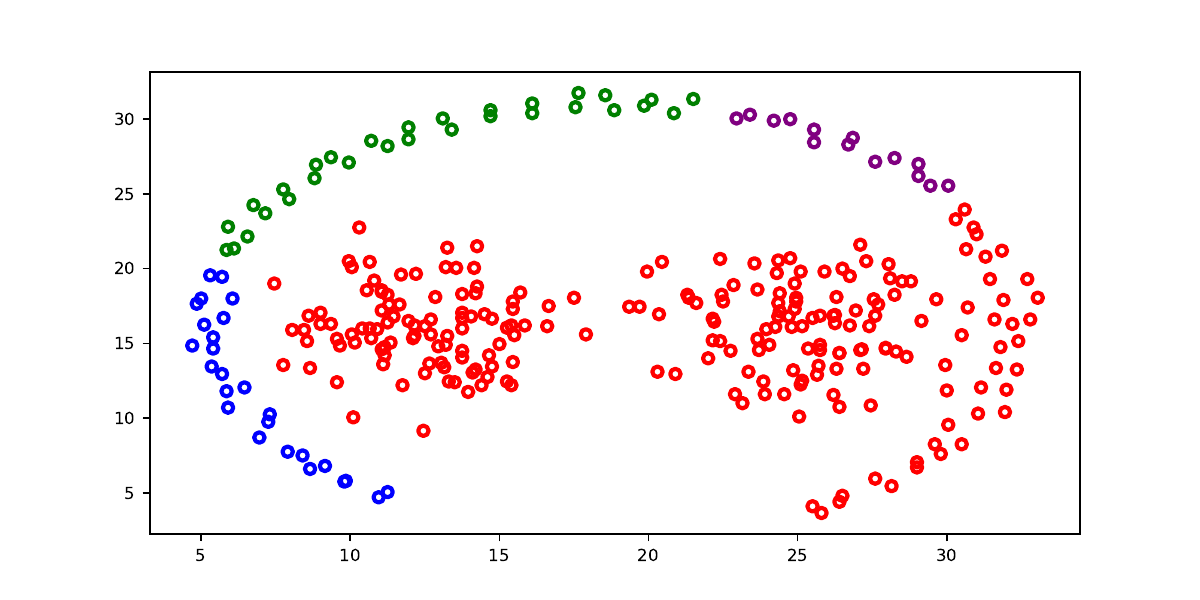} 
    \caption{Dunn index}
    \label{c}
  \end{subfigure}
  \hspace{-5.4mm}
  \begin{subfigure}[b]{0.45\textwidth}
    \centering
    \includegraphics[width=\textwidth]{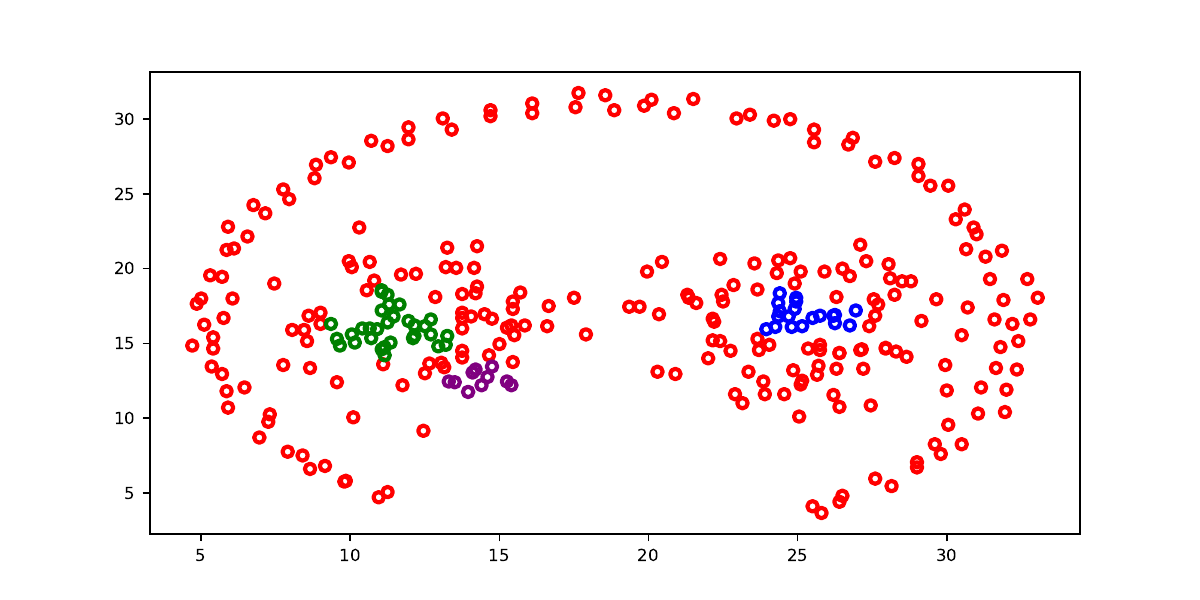} 
    \caption{Silhouette index}
    \label{fig:path_silhouette}
  \end{subfigure}
  \hspace{-5.4mm}
  \begin{subfigure}[b]{0.45\textwidth}
    \centering
    \includegraphics[width=\textwidth]{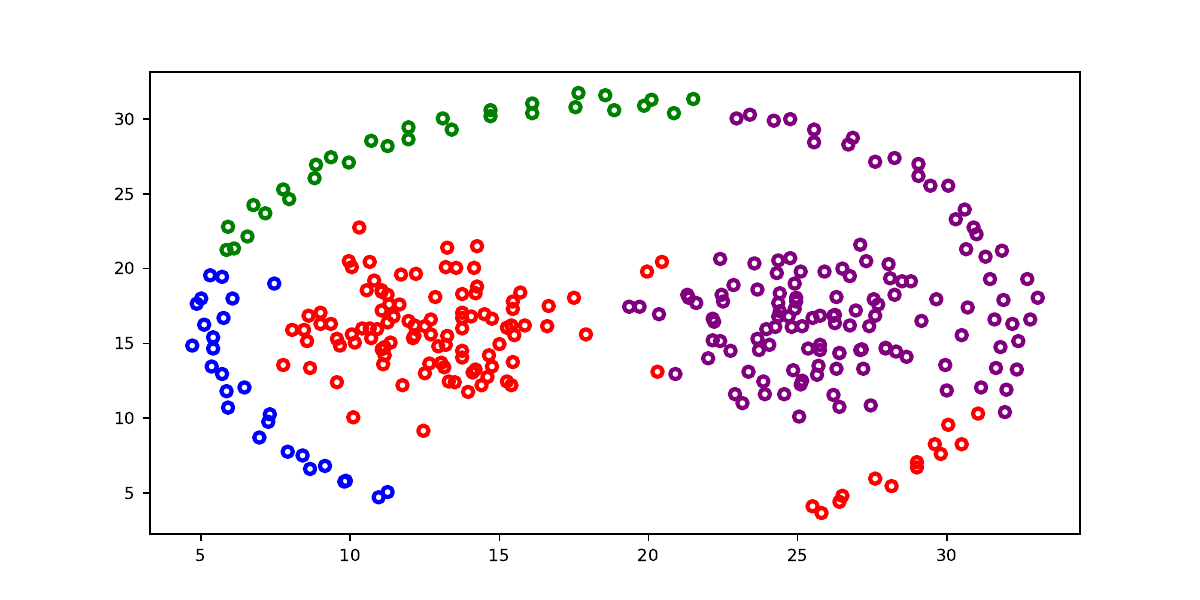} 
    \caption{XB index}
    \label{e}
  \end{subfigure}
    \begin{subfigure}[b]{0.45\textwidth}
    \centering
    \includegraphics[width=\textwidth]{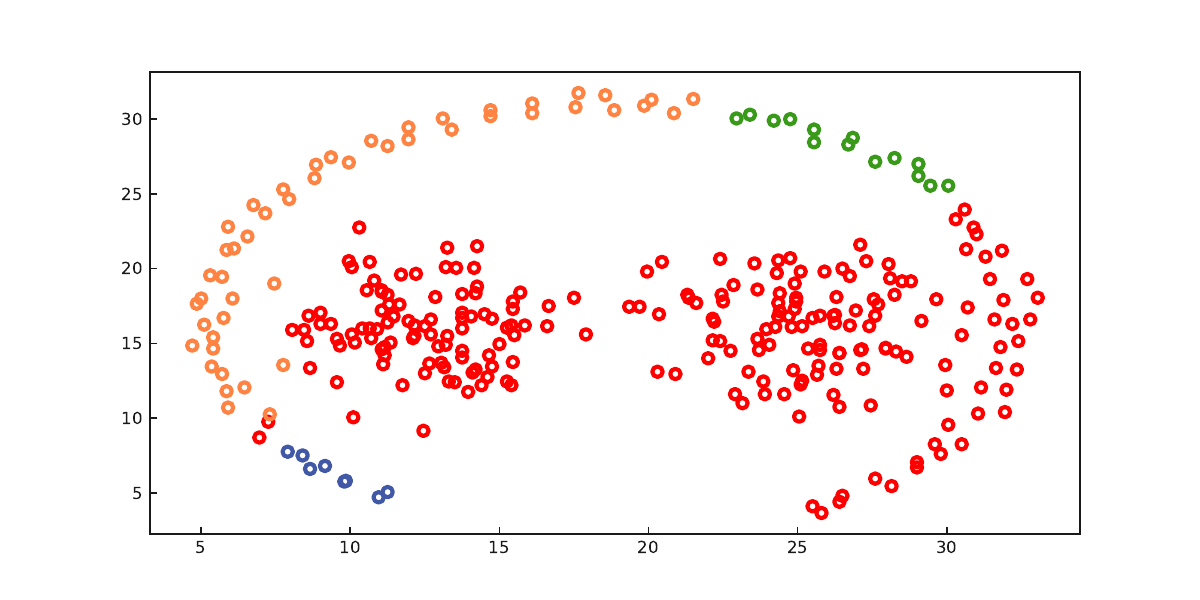} 
    \caption{CH index}
    \label{f}
  \end{subfigure}
  \hspace{-5.4mm}
  \begin{subfigure}[b]{0.45\textwidth}
    \centering
    \includegraphics[width=\textwidth]{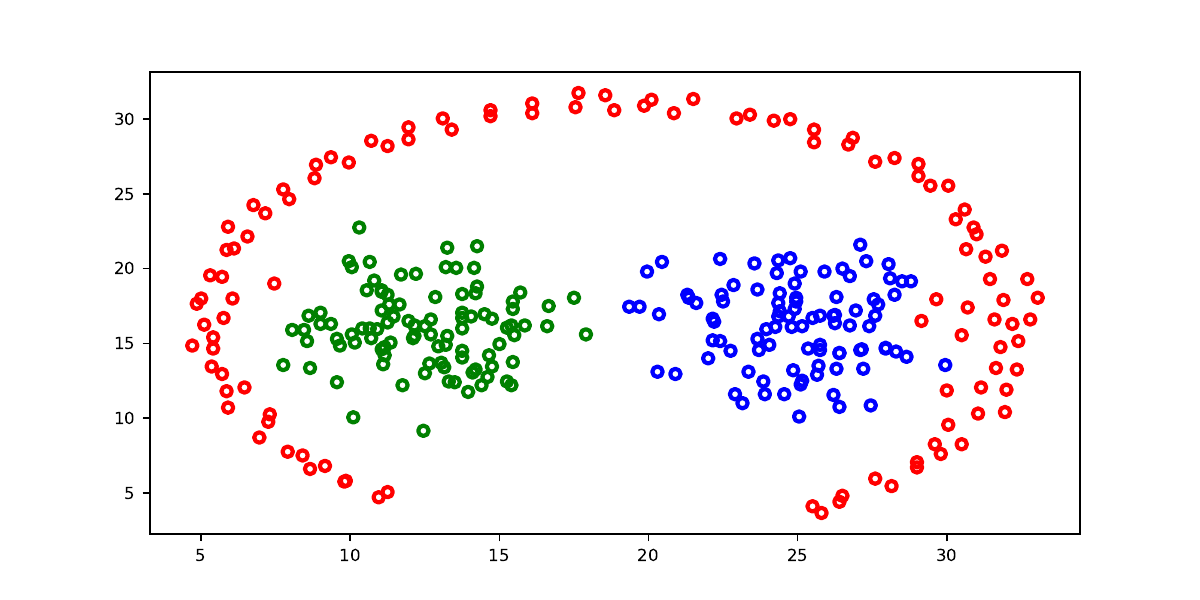} 
    \caption{Ground Truth}
    \label{fig:path_ground_truth}
  \end{subfigure}
  \hspace{-5.4mm}
  \caption{Path-based dataset with DBSCAN clustering and validation-index-based hyperparameter selection: (a) Proposed CDL index, (b) DB index, (c) Dunn index, (d) Silhouette index, (e) XB index, (f) CH index, and (g) ground truth.} \label{fig:all_for_path}
\end{figure}

For the arbitrary shape datasets we compare DBSCAN, OPTICS, and spectral clustering. The benchmarks are Path-Based, Aggregation, Jain's Toy, D31, and S15 \cite{lee2018}, shown in Figure~\ref{fig:2-sim}. Each dataset has its own distinctive cluster geometry: Path-Based has two symmetric branches, Aggregation has subclusters connected through a narrow neck, and the Path-Based and Jain's Toy datasets are not linearly separable. Figure~\ref{fig:all_for_path} shows the DBSCAN partitions obtained by selecting $\epsilon$ with each of the compared CVIs on Path-Based. For the three clustering methods (DBSCAN, OPTICS, and spectral clustering) CDL-CVI is compared with the Davies-Bouldin (DB), Dunn, Calinski-Harabasz (CH), Xie-Beni (XB), and Silhouette indices. ARI close to $1.0$ and NVI close to $0.0$ indicate full agreement with the reference partition; both are computed only after hyperparameter selection. Tables~\ref{Tab_1}, \ref{Tab_2}, and~\ref{Tab_3} report the cluster number selected by each validation index for the three clustering algorithms; entries are written as ``\textit{cluster number(ARI, NVI)}''. CDL-CVI selected the reference number of clusters on all listed arbitrary shape datasets and reached the highest ARI among the tested indices in the reported runs.

Traditional CVIs are sometimes paired with a kernel that maps the data to a higher dimensional space so that the clusters appear more compact and better separated. This requires the choice and tuning of both the kernel function and its parameters and is computationally expensive on large or high dimensional data. For a single kernel based reference, the tables include the Silhouette index combined with a kernel transformation, which gave the best results among the conventional methods we tested. With the kernel, the average processing time of DBSCAN rose from about 1 minute to at least 20 minutes, that of OPTICS from about 2 minutes to about 30 minutes, and that of spectral clustering from about 5 minutes to about 50 minutes. CDL-CVI does not require a kernel preprocessing step in these experiments.
CDL-CVI can also be used to compare clustering methods that share the same data and candidate hyperparameter grid. In our experiments the OPTICS partitions selected by CDL-CVI reach the highest ARI on every dataset, slightly above the DBSCAN and spectral partitions. The kernel based Silhouette baseline also tends to prefer OPTICS to DBSCAN and spectral clustering, but the resulting partitions reach lower ARI than those selected by CDL-CVI. The other CVIs (DB, Dunn, CH, XB, and Silhouette without kernel) are not as consistent across datasets and algorithms.

\begin{table}[]
\centering
\caption{Estimated number of clusters, ARI, and NVI for the DBSCAN algorithm (averaged over 100 trials)}
\label{Tab_1}
\resizebox{\textwidth}{!}{%
\begin{tabular}{|c|c|c|c|c|c|c|c|c|c|}
\hline
dataset &
  Size &
  \begin{tabular}[c]{@{}c@{}}cluster\\ number\end{tabular} &
  DB index &
  Dunn index &
  CH index &
  XB index &
  \begin{tabular}[c]{@{}c@{}}Silhouette\\ index\end{tabular} &
  \begin{tabular}[c]{@{}c@{}}Kernel with\\ silhouette index\end{tabular} &
  CDL-CVI \\ \hline
Path-based &
  300 &
  3 &
  2(0.6,0.2) &
  4(0.6,0.2) &
  4(0.6,0.3) &
  4(0.5,0.4) &
  4(0.6,0.3) &
  4(0.4, 0.1) &
  \textbf{3(0.81,0.14)} \\ \hline
Aggregation &
  788 &
  7 &
  6(0.9,0.2) &
  4(0.7,0.2) &
  6(0.7,0.3) &
  \textbf{7}(0.8,0.2) &
  3(0.7,0.3) &
  \textbf{7}(0.6,0.2) &
  \textbf{7(0.88,0.16)} \\ \hline
Jain's Toy &
  373 &
  2 &
  4(0.7,0.2) &
  \textbf {2}(0.8,0.1) &
  \textbf{2}(0.6,0.2) &
  1(0.8,0.2) &
  \textbf{2}(0.8,0.2) &
  4(0.5,0.1) &
  \textbf{2(0.89,0.13)} \\ \hline
D31 &
  3100 &
  31 &
  28(0.7,0.2) &
  27(0.6,0.3) &
  38(0.5,0.2) &
  21(0.6,0.2) &
  30(0.7,0.3) &
  35(0.7,0.1) &
  \textbf{31(0.90,0.18)} \\ \hline
S15 &
  600 &
  15 &
  12(0.7,0.2) &
  9(0.5,0.2) &
  15(0.7,0.2) &
  16(0.5,0.2) &
  30(0.6,0.4) &
  18(0.6,0.1) &
  \textbf{15(0.91,0.17)} \\ \hline
\end{tabular}%
}
\vspace{3mm}
\centering
\caption{Estimated number of clusters, ARI and NVI for the OPTICS algorithm (averaged over 100 trials)}
\label{Tab_2}
\resizebox{\textwidth}{!}{%
\begin{tabular}{|c|c|c|c|c|c|c|c|c|c|}
\hline
dataset &
  Size &
  \begin{tabular}[c]{@{}c@{}}cluster\\ number\end{tabular} &
  DB index &
  Dunn index &
  CH index &
  XB index &
  \begin{tabular}[c]{@{}c@{}}Silhouette\\ index\end{tabular} &
  \begin{tabular}[c]{@{}c@{}}Kernel with\\ silhouette index\end{tabular} &
  CDL-CVI \\ \hline
Path-based  & 300  & 3  & 2(0.7,0.2)  & 3(0.7,0.2)  & 2(0.7,0.4)  & 4(0.6,0.2)  & 2(0.8,0.2)          & \textbf{3(0.8, 0.1)} & \textbf{3(0.88,0.13)}  \\ \hline
Aggregation & 788  & 7  & 6(0.8,0.2)  & 5(0.6,0.3)  & 6(0.7,0.2)  & 6(0.7,0.3)  & 4(0.5,0.3)          & \textbf{7(0.9,0.1)}& \textbf{7(0.91,0.15)}  \\ \hline
Jain's Toy  & 373  & 2  & 4(0.7,0.2)  & 4(0.5,0.2)  & 3(0.6,0.3)  & 4(0.5,0.3)  & \textbf{2}(0.8,0.2) & 3(0.4, 0.1)         & \textbf{2(0.93,0.11)}  \\ \hline
D31         & 3100 & 31 & 28(0.5,0.1) & 27(0.6,0.3) & 33(0.6,0.3) & 27(0.5,0.4) & 30(0.8,0.2)         & 29(0.8, 0.1)       & \textbf{31(0.91,0.15)} \\ \hline
S15         & 600  & 15 & 7(0.5,0.3)  & 10(0.5,0.2) & 11(0.5,0.2) & 16(0.7,0.3) & 16(0.6,0.4)         & \textbf{15}(0.8,0.1) & \textbf{15(0.90,0.14)} \\ \hline
\end{tabular}%
}
\vspace{3mm}
\centering
\caption{Estimated number of clusters, ARI and NVI for the spectral clustering algorithm (averaged over 100 trials)}
\label{Tab_3}
\resizebox{\textwidth}{!}{%
\begin{tabular}{|c|c|c|c|c|c|c|c|c|c|}
\hline
dataset &
  Size &
  \begin{tabular}[c]{@{}c@{}}cluster\\ number\end{tabular} &
  DB index &
  Dunn index &
  CH index &
  XB index &
  \begin{tabular}[c]{@{}c@{}}Silhouette\\ index\end{tabular} &
  \begin{tabular}[c]{@{}c@{}}Kernel with\\ silhouette index\end{tabular} &
  CDL-CVI \\ \hline
Path-based  & 300  & 3  & 2(0.7,0.2)  & 3(0.7,0.2)  & 2(0.7,0.4)  & 4(0.6,0.2)  & 2(0.8,0.2)  & \textbf{3}(0.7, 0.1)  & \textbf{3(0.85,0.19)}  \\ \hline
Aggregation & 788  & 7  & 6(0.8,0.2)  & 5(0.6,0.3)  & 6(0.7,0.2)  & 6(0.7,0.3)  & 4(0.5,0.3)  & \textbf{7}(0.8,0.2)   & \textbf{7(0.88,0.19)}  \\ \hline
Jain's Toy  & 373  & 2  & 4(0.7,0.2)  & 4(0.5,0.2)  & 3(0.6,0.3)  & 4(0.5,0.3)  & 4(0.8,0.3)  & 4(0.6,0.2)            & \textbf{2(0.83,0.18)}  \\ \hline
D31         & 3100 & 31 & 28(0.5,0.1) & 27(0.6,0.3) & 35(0.7,0.3) & 28(0.6,0.4) & 33(0.8,0.2) & 28(0.6,0.1)           & \textbf{31(0.87,0.16)} \\ \hline
S15         & 600  & 15 & 9(0.6,0.4)  & 10(0.6,0.3) & 11(0.5,0.2) & 17(0.6,0.3) & 18(0.8,0.4) & 14(0.7,0.2)           & \textbf{15(0.88,0.17)} \\ \hline
\end{tabular}%
}
\end{table}

\subsection{Real Dataset}
This subsection evaluates CDL-CVI on a pipeline that is common in unsupervised engineering analytics: a frozen, pretrained representation produced by a deep model, followed by a classical clustering algorithm whose hyperparameters have to be chosen without labels. Similar pipelines arise in condition monitoring with pretrained signal embeddings, image based defect grouping, and exploratory analysis of measurement data, where the practitioner already has a domain relevant embedding but still needs an internal CVI to pick the number of clusters or the DBSCAN parameters. The benchmarks used here are MNIST (handwritten digits) \cite{lecun1998mnist}, CIFAR-10 \cite{krizhevsky2009learning}, and STL-10 (higher resolution natural images) \cite{coates2011analysis}.

MNIST digits are low resolution, centered, and reasonably well separated in pixel space, so classical algorithms work with appropriate preprocessing. CIFAR-10 (60k $32 \times 32$ colour images in 10 classes) has large intraclass variation, background clutter, and substantial feature overlap in raw pixel space; classical methods applied directly to pixels (or to a shallow linear reduction) recover only weak structure, internal CVIs tend to be close to zero, and the agreement with the ground truth is poor. A deep model is therefore typically used to extract features that are then passed to the classical clustering algorithm \cite{chen2019hybrid}. STL-10 contains 13k $96 \times 96$ colour images (10 classes) and was designed to encourage unsupervised representation learning before any downstream classifier or clustering. Classical clustering on the raw STL-10 pixels is hard because the features are very high dimensional. The deep embedding baseline of \cite{xie2016unsupervised} provides an unsupervised representation on which classical clustering becomes feasible, and indicates that without strong features the internal CVI values are small and the agreement with the ground truth is modest.

We use frozen pretrained models to extract features so that CDL-CVI is evaluated on meaningful representations rather than on raw pixel similarities. For MNIST we use the encoder of a pretrained convolutional Variational Autoencoder (VAE) \cite{mnistVAE}, which gives compact latent vectors that preserve digit identity. For CIFAR-10 we use a pretrained ResNet backbone trained with the SimCLR contrastive learning framework \cite{chen2020simple}. For STL-10 we use embeddings from a pretrained SimCLR model trained on the STL-10 unlabeled split \cite{chen2020big}, which yields 512 dimensional features. In every case the pretrained model is used as is, without any fine tuning. The resulting embeddings are passed to K-means, DBSCAN, and spectral clustering with Euclidean distance.

The hyperparameter grid search is treated as an embarrassingly parallel workload, with each pair of algorithm and hyperparameter, and each random seed, defining an independent task scheduled across CPU cores and, on clusters, via job arrays with dynamic load balancing. To reduce overhead, pairwise distance matrices and $k$ nearest neighbour graphs are precomputed and shared between candidate runs (for DBSCAN with a precomputed metric and for spectral clustering with precomputed affinities). For K-means, multiple random initializations are evaluated in parallel and the initialization with the smallest validation index value is kept.

The clustering results on the three real datasets are reported in Tables~\ref{tab:kmeans-table}, \ref{tab:dbscan-table}, and~\ref{tab:spectral-table}. Across all algorithms, CDL-CVI returns cluster numbers that are closest to the reference class counts in most cases and reaches the highest ARI among the tested indices in the reported trials. The same tables also show that CDL-CVI reaches the lowest NVI in most rows, so the gain in cluster quality is consistent between the two external metrics. NVI is reported with the convention that lower is better, so the bolded NVI entries are the smallest values in their row. For K-means, CDL-CVI recovers the reference number of clusters on MNIST and CIFAR-10, and on STL-10 it returns a cluster count closer to the reference than any other index; several of the other indices underestimate the cluster count. STL-10 remains a hard dataset and every method reaches only modest agreement with the ground truth, in line with what has been reported in the unsupervised representation learning literature \cite{xie2016unsupervised}. For DBSCAN, CDL-CVI again gives the best cluster estimates among the tested indices, whereas the Silhouette and Dunn indices often select an $\epsilon$ that is too large and underclusters the data. Because noise is excluded from both the internal score and the post selection ARI and NVI for DBSCAN, the comparison between the CVIs on DBSCAN is at parity, but the absolute DBSCAN numbers are not directly comparable to those of K-means and spectral clustering, which do not produce noise labels.

\begin{table}[]
\centering
\caption{K-means: estimated cluster numbers selected by each index, with ARI and NVI in parentheses.}
\label{tab:kmeans-table}
\resizebox{\textwidth}{!}{%
\begin{tabular}{|c|c|c|c|c|c|c|c|c|c|}
\hline
Dataset &
  Size &
  \begin{tabular}[c]{@{}c@{}}Cluster\\ Numbers\end{tabular} &
  \begin{tabular}[c]{@{}c@{}}Computation\\ Time (h:m)\end{tabular} &
  DB index &
  Dunn index &
  CH index &
  XB index &
  \begin{tabular}[c]{@{}c@{}}Silhouette\\ index\end{tabular} &
  CDL-CVI \\ \hline
MNIST          & 70{,}000 & 10 & 1:48 & 10(0.36, 0.75) &  7(0.33, 0.72) & 12(0.34, 0.73) & 10(0.37, 0.76) &  8(0.33, 0.73) & \textbf{10(0.41, 0.78)} \\ \hline
CIFAR-10       & 60{,}000 & 10 & 2:52 &  8(0.10, 0.56) &  5(0.08, 0.53) & 10(0.09, 0.54) &  9(0.11, 0.57) &  6(0.09, 0.55) & \textbf{10(0.14, 0.60)} \\ \hline
STL-10         & 13{,}000 & 10 & 0:57 &  7(0.07, 0.42) &  5(0.05, 0.40) &  8(0.06, 0.41) &  7(0.08, 0.43) &  6(0.06, 0.41) & \textbf{9(0.11, 0.47)} \\ \hline
\end{tabular}%
}
\end{table}

\begin{table}[]
\centering
\caption{DBSCAN: estimated cluster numbers selected by each index, with ARI and NVI in parentheses (noise excluded).}
\label{tab:dbscan-table}
\resizebox{\textwidth}{!}{%
\begin{tabular}{|c|c|c|c|c|c|c|c|c|c|}
\hline
Dataset &
  Size &
  \begin{tabular}[c]{@{}c@{}}Cluster\\ Numbers\end{tabular} &
  \begin{tabular}[c]{@{}c@{}}Computation\\ Time (h:m)\end{tabular} &
  DB index &
  Dunn index &
  CH index &
  XB index &
  \begin{tabular}[c]{@{}c@{}}Silhouette\\ index\end{tabular} &
  CDL-CVI \\ \hline
MNIST          & 70{,}000 & 10 & 4:05 &  9(0.37, 0.76) &  7(0.34, 0.73) & 11(0.35, 0.74) & 10(0.38, 0.77) &  8(0.35, 0.74) & \textbf{10(0.41, 0.80)} \\ \hline
CIFAR-10       & 60{,}000 & 10 & 6:12 &  6(0.09, 0.55) &  5(0.08, 0.53) &  8(0.08, 0.54) &  6(0.10, 0.56) &  5(0.08, 0.53) & \textbf{8(0.12, 0.58)} \\ \hline
STL-10         & 13{,}000 & 10 & 1:18 &  7(0.06, 0.40) &  5(0.05, 0.38) &  8(0.05, 0.39) &  7(0.07, 0.41) &  5(0.05, 0.39) & \textbf{7(0.09, 0.44)} \\ \hline
\end{tabular}%
}
\end{table}

\begin{table}[]
\centering
\caption{Spectral clustering: estimated cluster numbers selected by each index, with ARI and NVI in parentheses.}
\label{tab:spectral-table}
\resizebox{\textwidth}{!}{%
\begin{tabular}{|c|c|c|c|c|c|c|c|c|c|}
\hline
Dataset &
  Size &
  \begin{tabular}[c]{@{}c@{}}Cluster\\ Numbers\end{tabular} &
  \begin{tabular}[c]{@{}c@{}}Computation\\ Time (h:m)\end{tabular} &
  DB index &
  Dunn index &
  CH index &
  XB index &
  \begin{tabular}[c]{@{}c@{}}Silhouette\\ index\end{tabular} &
  CDL-CVI \\ \hline
MNIST          & 70{,}000 & 10 & 6:55 & 10(0.44, 0.79) &  8(0.41, 0.77) & 12(0.43, 0.78) & 10(0.45, 0.80) &  8(0.42, 0.78) & \textbf{10(0.49, 0.83)} \\ \hline
CIFAR-10       & 60{,}000 & 10 & 8:50 &  9(0.16, 0.60) &  6(0.14, 0.57) & 11(0.15, 0.59) &  9(0.17, 0.61) &  7(0.15, 0.58) & \textbf{10(0.20, 0.64)} \\ \hline
STL-10         & 13{,}000 & 10 & 2:10 &  7(0.10, 0.44) &  5(0.08, 0.42) &  8(0.09, 0.43) &  7(0.11, 0.45) &  6(0.09, 0.43) & \textbf{8(0.13, 0.48)} \\ \hline
\end{tabular}%
}
\end{table}

\section{Conclusion}
\label{sec:concl}
This paper introduced the Central Description Length Clustering Validation Index (CDL-CVI), an internal CVI that uses the observed cluster compactness and the estimated within cluster covariances to compute a probabilistic upper bound on the description length of the unobservable true cluster centers. The bound combines intra cluster compactness and centroid displacement into a single computable quantity, so candidate clusterings are ranked by one term instead of by two heterogeneous compactness and separation terms. The index does not use ground truth labels at run time, it is independent of the clustering algorithm that produced the candidate partition, and it applies to hyperparameters that fix the number of clusters implicitly (for example DBSCAN's $\epsilon$). On the synthetic three rings and arbitrary shape benchmarks (DBSCAN, OPTICS, and spectral clustering), CDL-CVI selected the reference number of clusters on every tested dataset and reached higher ARI than the conventional CVIs in our experiments, without any kernel preprocessing. On the image benchmarks (MNIST, CIFAR-10, STL-10) clustered from frozen unsupervised embeddings with K-means, DBSCAN, and spectral clustering, CDL-CVI returned cluster numbers close to the reference class counts and reached the highest ARI among the tested indices in the reported trials.

The pipeline used in the real dataset experiments, a frozen pretrained representation followed by a classical clustering algorithm whose hyperparameters are picked by an internal CVI, is common in engineering machine learning workflows where labels are scarce. A CVI that handles non convex cluster geometry, that works across clustering algorithms with the same code, and that does not require kernel tuning is of practical interest in such pipelines, whether the upstream representation is supervised, self supervised, or hand engineered.

The proposed bound has known assumptions and dependencies. The closed form expressions assume zero mean, finite variance, within cluster independent scatter, with a Gaussian approximation for the variance term. Deviations from these assumptions are absorbed asymptotically by central limit arguments, but the bound becomes less tight on small clusters, on heavy tailed data, and in strongly anisotropic high dimensional regimes where covariance estimation is unstable. The bound also depends on the distance metric (Euclidean in this paper) and on feature scaling, and should be applied after a problem appropriate normalization. Like every internal CVI evaluated by grid search, CDL-CVI inherits the bias of the search grid; our results therefore describe the behaviour on the candidate clusterings produced by the grids we used, rather than the behaviour of the clustering algorithm in isolation. The Chebyshev bound is also conservative; tighter sub Gaussian or Bernstein style concentration results, when their assumptions can be justified, would likely improve it, especially in the low sample regime.

Natural extensions of this work include robust covariance estimation for high dimensional embeddings, density and graph aware variants in which the central error is defined relative to local centers or representatives rather than Euclidean means, and evaluation on engineering datasets such as vibration and acoustic measurements for fault diagnosis, sensor state discovery in industrial processes, structural health monitoring, and materials characterization. Statistical uncertainty quantification of the selected hyperparameter (for example via bootstrap over the dataset and over the candidate grid) and tighter concentration analyses are other natural directions.
%
%




\data{The benchmark datasets analysed in this study are publicly available from the sources cited in the manuscript. The three rings dataset is generated synthetically from the parameters given in Section~\ref{seq:sim}. The arbitrary shape benchmarks (Aggregation, D31, Jain's Toy, Path-Based, S15) are taken from the public repository associated with \cite{lee2018}, and MNIST \cite{lecun1998mnist}, CIFAR-10 \cite{krizhevsky2009learning}, and STL-10 \cite{coates2011analysis} are obtained from their official distributions. The pretrained embedding models used for the real data experiments (the MNIST VAE \cite{mnistVAE} and the SimCLR and ResNet backbones \cite{chen2020simple, chen2020big}) are publicly available from the cited sources. A reference implementation of CDL-CVI, the hyperparameter grids, random seeds, and the scripts that produce all tables and figures will be released in a public repository upon acceptance. No new experimental dataset was created for this study.}


\section*{References}
\bibliographystyle{iopart-num-titles}
\bibliography{refs}

@article{beheshti2020k,
  title={{K-MACE} and Kernel {K-MACE} Clustering},
  author={Beheshti, Soosan and Nidoy, Edward and Rahman, Faizan},
  journal={IEEE Access},
  volume={8},
  pages={17390--17403},
  year={2020},
  publisher={IEEE}
}

@article{arbelaitz2013,
  title={An extensive comparative study of cluster validity indices},
  author={Arbelaitz, Olatz and Gurrutxaga, Ibai and Muguerza, Javier}, 
  journal={Pattern Recognition},
  volume={46},
  number={1},
  pages={243--256},
  year={2013},
  publisher={Elsevier}
}

@article{lee2018,
  title={A new clustering validity index for arbitrary shape of clusters},
  author={Lee, Soo-Hyun and Jeong, Young-Seon and Kim, Jae-Yun and Jeong, Myong K},
  journal={Pattern Recognition Letters},
  volume={112},
  pages={263--269},
  year={2018},
  publisher={Elsevier}
}

@article{jain2010,
  title={Data clustering: 50 years beyond {K}-means},
  author={Jain, Anil K},
  journal={Pattern recognition letters},
  volume={31},
  number={8},
  pages={651--666},
  year={2010},
  publisher={Elsevier}
}

@article{gurrutxaga2011,
  title={Towards a standard methodology to evaluate internal cluster validity indices},
  author={Gurrutxaga, Ibai and Muguerza, Javier and Arbelaitz, Olatz and P{\'e}rez, Jes{\'u}s M and Mart{\'\i}n, Jos{\'e} I},
  journal={Pattern Recognition Letters},
  volume={32},
  number={3},
  pages={505--515},
  year={2011},
  publisher={Elsevier}
}

@article{dunn1973,
  title={A fuzzy relative of the {ISODATA} process and its use in detecting compact well-separated clusters},
  author={Dunn, Joseph C},
  journal={Journal of Cybernetics},
  volume={3},
  number={3},
  pages={32--57},
  year={1973},
  publisher={Taylor \& Francis},
  doi={10.1080/01969727308546046}
}

@article{calinski1974,
  title={A dendrite method for cluster analysis},
  author={Cali{\'n}ski, Tadeusz and Harabasz, Jerzy},
  journal={Communications in Statistics-theory and Methods},
  volume={3},
  number={1},
  pages={1--27},
  year={1974},
  publisher={Taylor \& Francis}
}

@article{davies1979,
  title={A cluster separation measure},
  author={Davies, David L and Bouldin, Donald W},
  journal={IEEE Transactions on Pattern Analysis and Machine Intelligence},
  volume={PAMI-1},
  number={2},
  pages={224--227},
  year={1979},
  publisher={IEEE},
  doi={10.1109/TPAMI.1979.4766909}
}

@article{rousseeuw1987silhouettes,
  title={Silhouettes: a graphical aid to the interpretation and validation of cluster analysis},
  author={Rousseeuw, Peter J},
  journal={Journal of computational and applied mathematics},
  volume={20},
  pages={53--65},
  year={1987},
  publisher={Elsevier}
}

@article{shahbaba2014,
  title={{MACE}-means clustering},
  author={Shahbaba, Mahdi and Beheshti, Soosan},
  journal={Signal Processing},
  volume={105},
  pages={216--225},
  year={2014},
  publisher={Elsevier}
}

@inproceedings{rahman2018,
  title={Kernel {K-MACE} Clustering},
  author={Rahman, Faizan and Beheshti, Soosan},
  booktitle={2018 52nd Asilomar Conference on Signals, Systems, and Computers},
  pages={2002--2006},
  year={2018},
  organization={IEEE}
}

@article{Beheshti2005,
  title={A new information-theoretic approach to signal denoising and best basis selection},
  author={Beheshti, Soosan and Dahleh, Munther A},
  journal={IEEE Transactions on Signal Processing},
  volume={53},
  number={10},
  pages={3613--3624},
  year={2005},
  publisher={IEEE}
}

@inproceedings{shamsi2019correct,
  title={Correct number of clusters ({CNC}) description length in arbitrary shape clustering},
  author={Shamsi, Mahdi and Rahman, Faizan and Beheshti, Soosan},
  booktitle={2019 16th Canadian Workshop on Information Theory (CWIT)},
  pages={1--4},
  year={2019},
  organization={IEEE}
}

@book{grunwald2007minimum,
  title={The minimum description length principle},
  author={Gr{\"u}nwald, Peter D},
  year={2007},
  publisher={MIT press}
}

@inproceedings{fakhrzadeh2007minimum,
  title={Minimum noiseless description length ({MNDL}) thresholding},
  author={Fakhrzadeh, Azadeh and Beheshti, Soosan},
  booktitle={2007 IEEE Symposium on Computational Intelligence in Image and Signal Processing},
  pages={146--150},
  year={2007},
  organization={IEEE}
}

@article{motallebi2022local,
  title={A local mean-based distance measure for spectral clustering},
  author={Motallebi, Hassan and Nasihatkon, Rabeeh and Jamshidi, Mina},
  journal={Pattern Analysis and Applications},
  volume={25},
  number={2},
  pages={351--359},
  year={2022},
  publisher={Springer}
}

@inproceedings{liu2010understanding,
  title={Understanding of internal clustering validation measures},
  author={Liu, Yanchi and Li, Zhongmou and Xiong, Hui and Gao, Xuedong and Wu, Junjie},
  booktitle={2010 IEEE international conference on data mining},
  pages={911--916},
  year={2010},
  organization={IEEE}
}

@inproceedings{khan2014dbscan,
  title={{DBSCAN}: Past, present and future},
  author={Khan, Kamran and Rehman, Saif Ur and Aziz, Kamran and Fong, Simon and Sarasvady, Sababady},
  booktitle={The fifth international conference on the applications of digital information and web technologies (ICADIWT 2014)},
  pages={232--238},
  year={2014},
  organization={IEEE}
}

@article{xie1991validity,
  title={A validity measure for fuzzy clustering},
  author={Xie, Xuanli Lisa and Beni, Gerardo},
  journal={IEEE Transactions on Pattern Analysis \& Machine Intelligence},
  volume={13},
  number={08},
  pages={841--847},
  year={1991},
  publisher={IEEE Computer Society}
}

@book{cover1999elements,
  title={Elements of information theory},
  author={Cover, Thomas M},
  year={1999},
  publisher={John Wiley \& Sons}
}

@book{rodrigues2021information,
  title={Information-theoretic methods in data science},
  author={Rodrigues, Miguel RD and Eldar, Yonina C},
  year={2021},
  publisher={Cambridge University Press}
}

@article{bandyapadhyayfind,
  title={How to find a good explanation for clustering?},
  author={Bandyapadhyay, Sayan and Fomin, Fedor V and Golovach, Petr A and Lochet, William and Purohit, Nidhi and Simonov, Kirill},
  journal={Artificial Intelligence},
  volume={322},
  pages={103948},
  year={2023},
  publisher={Elsevier}
}

@article{li2019clustering,
  title={Clustering ensemble based on sample's stability},
  author={Li, Feijiang and Qian, Yuhua and Wang, Jieting and Dang, Chuangyin and Jing, Liping},
  journal={Artificial Intelligence},
  volume={273},
  pages={37--55},
  year={2019},
  publisher={Elsevier}
}

@article{BeheshtiEcoac2025,
  title={$\epsilon$-{C}onfidence {A}pproximately {C}orrect ($\epsilon$-{CoAC}) Learnability and Hyperparameter Selection in Linear Regression Modeling},
  author={Beheshti, Soosan and Shamsi, Mahdi},
  journal={IEEE Access},
  year={2025},
  note={in press},
  publisher={IEEE}
}

@misc{lecun1998mnist,
  title={The {MNIST} database of handwritten digits},
  author={LeCun, Yann and Cortes, Corinna and Burges, Christopher J C},
  year={1998},
  howpublished={\url{http://yann.lecun.com/exdb/mnist/}},
  note={Accessed: 2025-08-22}
}

@inproceedings{coates2011analysis,
  title={An analysis of single-layer networks in unsupervised feature learning},
  author={Coates, Adam and Ng, Andrew and Lee, Honglak},
  booktitle={Proceedings of the fourteenth international conference on artificial intelligence and statistics},
  pages={215--223},
  year={2011},
  organization={JMLR Workshop and Conference Proceedings}
}

@techreport{krizhevsky2009learning,
  title={Learning multiple layers of features from tiny images},
  author={Krizhevsky, Alex and Hinton, Geoffrey},
  year={2009},
  institution={University of Toronto},
  type={Technical Report}
}

@article{chen2019hybrid,
  title={A hybrid autoencoder network for unsupervised image clustering},
  author={Chen, Pei-Yin and Huang, Jih-Jeng},
  journal={Algorithms},
  volume={12},
  number={6},
  pages={122},
  year={2019},
  publisher={MDPI}
}

@inproceedings{xie2016unsupervised,
  title={Unsupervised deep embedding for clustering analysis},
  author={Xie, Junyuan and Girshick, Ross and Farhadi, Ali},
  booktitle={International conference on machine learning},
  pages={478--487},
  year={2016},
  organization={PMLR}
}

@inproceedings{moulavi2014density,
  title={Density-based clustering validation},
  author={Moulavi, Davoud and Jaskowiak, Pablo A and Campello, Ricardo JGB and Zimek, Arthur and Sander, J{\"o}rg},
  booktitle={Proceedings of the 2014 SIAM international conference on data mining},
  pages={839--847},
  year={2014},
  organization={SIAM}
}

@article{cheng2018novel,
  title={A novel cluster validity index based on local cores},
  author={Cheng, Dongdong and Zhu, Qingsheng and Huang, Jinlong and Wu, Quanwang and Yang, Lijun},
  journal={IEEE transactions on neural networks and learning systems},
  volume={30},
  number={4},
  pages={985--999},
  year={2018},
  publisher={IEEE}
}

@article{hassan2024z,
  title={From {A}-to-{Z} review of clustering validation indices},
  author={Hassan, Bryar A and Tayfor, Noor Bahjat and Hassan, Alla A and Ahmed, Aram M and Rashid, Tarik A and Abdalla, Naz N},
  journal={Neurocomputing},
  volume={601},
  pages={128198},
  year={2024},
  publisher={Elsevier}
}

@article{diallo2023auto,
  title={Auto-attention mechanism for multi-view deep embedding clustering},
  author={Diallo, Bassoma and Hu, Jie and Li, Tianrui and Khan, Ghufran Ahmad and Liang, Xinyan and Wang, Hongjun},
  journal={Pattern Recognition},
  volume={143},
  pages={109764},
  year={2023},
  publisher={Elsevier}
}

@inproceedings{li2021contrastive,
  title={Contrastive clustering},
  author={Li, Yunfan and Hu, Peng and Liu, Zitao and Peng, Dezhong and Zhou, Joey Tianyi and Peng, Xi},
  booktitle={Proceedings of the AAAI conference on artificial intelligence},
  volume={35},
  pages={8547--8555},
  year={2021}
}

@misc{mnistVAE,
  author = {Sarasti, Sebasti{\'a}n},
  title = {{MNISTVae}},
  year = {2024},
  howpublished = {\url{https://huggingface.co/sebastiansarasti/MNISTVae}},
  note = {Accessed: 2025-08-22}
}

@inproceedings{chen2020simple,
  title={A simple framework for contrastive learning of visual representations},
  author={Chen, Ting and Kornblith, Simon and Norouzi, Mohammad and Hinton, Geoffrey},
  booktitle={International conference on machine learning},
  pages={1597--1607},
  year={2020},
  organization={PmLR}
}

@article{chen2020big,
  title={{Big Self-Supervised Models are Strong Semi-Supervised Learners}},
  author={Chen, Ting and Kornblith, Simon and Swersky, Kevin and Norouzi, Mohammad and Hinton, Geoffrey},
  journal={arXiv preprint arXiv:2006.10029},
  year={2020}
}

@article{khan2024complementary,
  title={Complementary incomplete weighted concept factorization methods for multi-view clustering},
  author={Khan, Ghufran Ahmad and Khan, Jalaluddin and Anwar, Taushif and Al-Huda, Zaid and Diallo, Bassoma and Ahmad, Naved},
  journal={Knowledge and Information Systems},
  volume={66},
  number={12},
  pages={7469--7494},
  year={2024},
  publisher={Springer}
}

@article{khan2023multi,
  title={Multi-view subspace clustering for learning joint representation via low-rank sparse representation},
  author={Khan, Ghufran Ahmad and Hu, Jie and Li, Tianrui and Diallo, Bassoma and Du, Shengdong},
  journal={Applied Intelligence},
  volume={53},
  number={19},
  pages={22511--22530},
  year={2023},
  publisher={Springer Nature BV}
}

@ARTICLE{saba,
  author={Beheshti, Soosan and Sedghizadeh, Saba},
  journal={IEEE Transactions on Signal Processing}, 
  title={Number of Source Signal Estimation by the Mean Squared Eigenvalue Error}, 
  year={2018},
  volume={66},
  number={21},
  pages={5694-5704},
  doi={10.1109/TSP.2018.2870357}}

@ARTICLE{tdelay,
  author={Shamsi, Mahdi and Beheshti, Soosan},
  journal={IEEE Access}, 
  title={Relative Entropy ({RE})-Based {LTI} System Modeling Equipped With Simultaneous Time Delay Estimation and Online Modeling},
  year={2023},
  volume={11},
  number={},
  pages={113885-113899},
  doi={10.1109/ACCESS.2023.3321794}}

@ARTICLE{vedant,
  author={Beheshti, Soosan and Bommanahally, Vedant},
  journal={IEEE Access}, 
  title={Minimum Mismatch Modeling ({3M}) Hyperparameter Selection in Autoregressive Moving Average ({ARMA}) Modeling},
  year={2025},
  volume={13},
  number={},
  pages={133681-133693},
  doi={10.1109/ACCESS.2025.3592088}}

@article{subid,
title = {Reliable truncation parameter selection and model order estimation for stochastic subspace identification},
journal = {Journal of the Franklin Institute},
pages = {107766},
year = {2025},
issn = {0016-0032},
doi = {https://doi.org/10.1016/j.jfranklin.2025.107766},
author = {Khashayar Bayati and Karthikeyan Umapathy and Soosan Beheshti},

}

\appendix

\section{Derivation of CDL-CVI Bounds}
\label{app:derivation}

This appendix summarizes the derivation that relates the observable cluster compactness $y_{K_\Theta}$ to the unobservable Average Central Error $z_{K_\Theta}$. The full algebra of the variance expressions is given in \cite{beheshti2020k}; the notation here follows the CDL-CVI formulation used in this paper. Throughout, the scatter factors $\omega_{x(i)}$ are assumed to be zero mean with finite second moments and mutually independent across the data points of a given estimated cluster. The Gaussian assumption is used only for the closed form variance expression in Lemma~2 and is replaced asymptotically by central limit arguments when the scatter is non Gaussian.

For a given clustering with $K_\Theta$ clusters, for each cluster $c_{K_\Theta}(k)$, we use matrix notation for the data points, their true centers, and their scatter factors.
\begin{itemize}
    \item $\mathbf{x}_k$: A $d \times n_{K_\Theta k}$ matrix where columns are the data points $x_{K_\Theta k}(i)$ in cluster $k$.
    \item $\overline{\boldsymbol{\mu}}_k$: A $d \times n_{K_\Theta k}$ matrix of the corresponding true centers $\overline{\mu}_{x_{K_\Theta k}(i)}$.
    \item $\boldsymbol{\omega}_k$: A $d \times n_{K_\Theta k}$ matrix of the corresponding scatter factors $\omega_{x_{K_\Theta k}(i)}$.
\end{itemize}
So, $\mathbf{x}_k = \overline{\boldsymbol{\mu}}_k + \boldsymbol{\omega}_k$.

The estimated center $\widehat{\mu}_{K_\Theta}(k)$ is the average of the columns of $\mathbf{x}_k$. This can be written as $\widehat{\mu}_{K_\Theta}(k) = \frac{1}{n_{K_\Theta k}} \mathbf{x}_k \mathbf{1}$, where $\mathbf{1}$ is a vector of ones of size $n_{K_\Theta k}$. To express the central error and cluster compactness in matrix form, we define two $n_{K_\Theta k} \times n_{K_\Theta k}$ matrices:
\begin{itemize}
    \item $B_{K_\Theta k}$: An averaging matrix where every element is $1/n_{K_\Theta k}$.
    \item $A_{K_\Theta k} = I - B_{K_\Theta k}$, where $I$ is the identity matrix.
\end{itemize}
Note that $A_{K_\Theta k}$ is symmetric and idempotent ($A_{K_\Theta k}^T A_{K_\Theta k} = A_{K_\Theta k}$), and $A_{K_\Theta k}^T B_{K_\Theta k} = 0$.

Using this notation, the cluster compactness for cluster $k$ (sum of squared distances from the estimated center) is the squared Frobenius norm:
\begin{align}
    y_{K_\Theta k} = \left\| \mathbf{x}_k - \widehat{\boldsymbol{\mu}}_k \right\|_F^2 = \left\| \mathbf{x}_k A_{K_\Theta k} \right\|_F^2
\end{align}
And the central error for cluster $k$ is:
\begin{align}
    z_{K_\Theta k} = \left\| \overline{\boldsymbol{\mu}}_k - \widehat{\boldsymbol{\mu}}_k \right\|_F^2 = \left\| \overline{\boldsymbol{\mu}}_k - \mathbf{x}_k B_{K_\Theta k} \right\|_F^2
\end{align}
where $\widehat{\boldsymbol{\mu}}_k$ is a $d \times n_{K_\Theta k}$ matrix with each column being the estimated center $\widehat{\mu}_{K_\Theta}(k)$.

\subsection[Mean and Variance of Central Error]{Mean and Variance of Central Error ($Z_{K_\Theta k}$)}

\begin{lemma}
The central error for the $k^{th}$ cluster, $z_{K_\Theta k}$, is a sample of a random variable $Z_{K_\Theta k}$. The expected value and variance of $Z_{K_\Theta k}$ are:
\begin{align}
    E[Z_{K_\Theta k}] &= \left\| \Delta_{K_\Theta k} \right\|_F^2 + \frac{1}{n_{K_\Theta k}} \sum_{i=1}^{n_{K_\Theta k}} \mathrm{tr}(\overline{\Lambda}_{x_{K_\Theta k}(i)}) \\
    \mathrm{Var}[Z_{K_\Theta k}] &= \frac{2}{n_{K_\Theta k}^2} \sum_{i=1}^{n_{K_\Theta k}} \mathrm{tr}((\overline{\Lambda}_{x_{K_\Theta k}(i)})^2) + \frac{2}{n_{K_\Theta k}^2} \sum_{i \neq j}^{n_{K_\Theta k}} \mathrm{tr}(\overline{\Lambda}_{x_{K_\Theta k}(i)} \overline{\Lambda}_{x_{K_\Theta k}(j)})
\end{align}
where $\left\| \Delta_{K_\Theta k} \right\|_F^2 = \left\| \overline{\boldsymbol{\mu}}_k A_{K_\Theta k} \right\|_F^2$ represents the variation of the true centers within the estimated cluster, and $\overline{\Lambda}_{x_{K_\Theta k}(i)}$ is the covariance matrix of the scatter factor $\omega_{x_{K_\Theta k}(i)}$.
\end{lemma}

Substitute $\mathbf{x}_k = \overline{\boldsymbol{\mu}}_k + \boldsymbol{\omega}_k$ into the definition of central error:
\begin{align}
    Z_{K_\Theta k} &= \left\| \overline{\boldsymbol{\mu}}_k - (\overline{\boldsymbol{\mu}}_k + \boldsymbol{\omega}_k) B_{K_\Theta k} \right\|_F^2 \\
    &= \left\| \overline{\boldsymbol{\mu}}_k (I - B_{K_\Theta k}) - \boldsymbol{\omega}_k B_{K_\Theta k} \right\|_F^2 \\
    &= \left\| \overline{\boldsymbol{\mu}}_k A_{K_\Theta k} - \boldsymbol{\omega}_k B_{K_\Theta k} \right\|_F^2 \\
    &= \left\| \overline{\boldsymbol{\mu}}_k A_{K_\Theta k} \right\|_F^2 + \left\| \boldsymbol{\omega}_k B_{K_\Theta k} \right\|_F^2 - 2\,\mathrm{tr}\!\left(B_{K_\Theta k}^T \boldsymbol{\omega}_k^T \overline{\boldsymbol{\mu}}_k A_{K_\Theta k}\right).
\end{align}
The cross term in the last line is linear in $\boldsymbol{\omega}_k$, so under $E[\boldsymbol{\omega}_k] = 0$ its expectation vanishes regardless of the structure of $\overline{\boldsymbol{\mu}}_k$, $A_{K_\Theta k}$, and $B_{K_\Theta k}$. The expected central error is then
\begin{align}
    E[Z_{K_\Theta k}] = \left\| \overline{\boldsymbol{\mu}}_k A_{K_\Theta k} \right\|_F^2 + E\left[\left\| \boldsymbol{\omega}_k B_{K_\Theta k} \right\|_F^2\right] = \left\| \Delta_{K_\Theta k} \right\|_F^2 + E\left[\left\| \boldsymbol{\omega}_k B_{K_\Theta k} \right\|_F^2\right]
\end{align}
The term $\left\| \boldsymbol{\omega}_k B_{K_\Theta k} \right\|_F^2$ can be expanded using the independence of the scatter factors $\omega(i)$ and $\omega(j)$ for $i \neq j$:
\begin{align}
    E[\left\| \boldsymbol{\omega}_k B_{K_\Theta k} \right\|_F^2] &= E[\mathrm{tr}(B_{K_\Theta k}^T \boldsymbol{\omega}_k^T \boldsymbol{\omega}_k B_{K_\Theta k})] \\
    &= \frac{1}{n_{K_\Theta k}} \sum_{i=1}^{n_{K_\Theta k}} E[\omega_{x_{K_\Theta k}(i)}^T \omega_{x_{K_\Theta k}(i)}] = \frac{1}{n_{K_\Theta k}} \sum_{i=1}^{n_{K_\Theta k}} \mathrm{tr}(\overline{\Lambda}_{x_{K_\Theta k}(i)})
\end{align}
The variance calculation follows similarly from the properties of the trace and the independence of the scatter factors, as detailed in \cite{beheshti2020k}.

\subsection[Mean and Variance of Cluster Compactness]{Mean and Variance of Cluster Compactness ($Y_{K_\Theta k}$)}

\begin{lemma}
The cluster compactness for the $k^{th}$ cluster, $y_{K_\Theta k}$, is a sample of a random variable $Y_{K_\Theta k}$. Under the zero-mean, mutually independent scatter assumption with covariance $\overline{\Lambda}_{x_{K_\Theta k}(i)}$, the expected value and variance admit the closed forms:
\begin{align}
    E[Y_{K_\Theta k}] &= \left\| \Delta_{K_\Theta k} \right\|_F^2 + \frac{n_{K_\Theta k}-1}{n_{K_\Theta k}} \sum_{i=1}^{n_{K_\Theta k}} \mathrm{tr}(\overline{\Lambda}_{x_{K_\Theta k}(i)}) \\
    \mathrm{Var}[Y_{K_\Theta k}] &= \frac{2(n_{K_\Theta k}-1)^2}{n_{K_\Theta k}^2} \sum_{i=1}^{n_{K_\Theta k}} \mathrm{tr}\!\left((\overline{\Lambda}_{x_{K_\Theta k}(i)})^2\right) \nonumber \\
    &\quad + \frac{2}{n_{K_\Theta k}^2}\sum_{i\neq j}\mathrm{tr}\!\left(\overline{\Lambda}_{x_{K_\Theta k}(i)}\overline{\Lambda}_{x_{K_\Theta k}(j)}\right)\nonumber\\
    &\quad + 4\,\mathrm{tr}\!\left(A_{K_\Theta k}\,\overline{\boldsymbol{\mu}}_k^T\,\overline{\boldsymbol{\mu}}_k\,A_{K_\Theta k}\,\Sigma_{\omega,k}\right),
\end{align}
where $\Sigma_{\omega,k}$ is the average per-sample scatter covariance in cluster $k$. The closed form for the variance is obtained under the Gaussian-scatter assumption; under non-Gaussian scatter the same expression holds asymptotically by central-limit-style arguments. The full algebra is given in Appendix~B of \cite{beheshti2020k}.
\end{lemma}

Substitute $\mathbf{x}_k = \overline{\boldsymbol{\mu}}_k + \boldsymbol{\omega}_k$ into the definition of cluster compactness:
\begin{align}
    Y_{K_\Theta k} &= \left\| (\overline{\boldsymbol{\mu}}_k + \boldsymbol{\omega}_k) A_{K_\Theta k} \right\|_F^2 \\
    &= \left\| \overline{\boldsymbol{\mu}}_k A_{K_\Theta k} + \boldsymbol{\omega}_k A_{K_\Theta k} \right\|_F^2 \\
    &= \left\| \Delta_{K_\Theta k} \right\|_F^2 + \left\| \boldsymbol{\omega}_k A_{K_\Theta k} \right\|_F^2 + 2\,\mathrm{tr}\!\left(A_{K_\Theta k}^T \boldsymbol{\omega}_k^T \overline{\boldsymbol{\mu}}_k A_{K_\Theta k}\right).
\end{align}
The cross term is linear in $\boldsymbol{\omega}_k$ and therefore has zero expectation under $E[\boldsymbol{\omega}_k] = 0$. The expectation of $\left\| \boldsymbol{\omega}_k A_{K_\Theta k} \right\|_F^2$ is
\begin{align}
    E[\left\| \boldsymbol{\omega}_k A_{K_\Theta k} \right\|_F^2] &= E[\mathrm{tr}(A_{K_\Theta k}^T \boldsymbol{\omega}_k^T \boldsymbol{\omega}_k A_{K_\Theta k})] \\
    &= \mathrm{tr}(A_{K_\Theta k} E[\boldsymbol{\omega}_k^T \boldsymbol{\omega}_k]) \\
    &= \frac{n_{K_\Theta k}-1}{n_{K_\Theta k}} \sum_{i=1}^{n_{K_\Theta k}} \mathrm{tr}(\overline{\Lambda}_{x_{K_\Theta k}(i)})
\end{align}
This gives the expression for $E[Y_{K_\Theta k}]$. The variance expression in Lemma~2 is obtained by computing the fourth order moment $E\!\left[(Y_{K_\Theta k} - E[Y_{K_\Theta k}])^2\right]$ via the Isserlis identity for Gaussian scatter. The full derivation is given in Appendix~B of \cite{beheshti2020k}.

\subsection[Probabilistic Bounds on Delta]{Probabilistic Bounds on $\left\| \Delta_{K_\Theta k} \right\|_F^2$}

Both $E[Z_{K_\Theta k}]$ and $E[Y_{K_\Theta k}]$ contain the same unknown term $\left\| \Delta_{K_\Theta k} \right\|_F^2$, so the observed cluster compactness $y_{K_\Theta k}$ can be used to bound this term probabilistically.

\begin{theorem}
With a validation probability $P_v = 1 - 1/\alpha_k^2$, the term $\left\| \Delta_{K_\Theta k} \right\|_F^2$ is bounded.
\end{theorem}

Using Chebyshev's inequality on the random variable $Y_{K_\Theta k}$:
\begin{align}
    P(|Y_{K_\Theta k} - E[Y_{K_\Theta k}]| \leq \alpha_k \sqrt{\mathrm{Var}[Y_{K_\Theta k}]}) \geq 1 - \frac{1}{\alpha_k^2}
\end{align}
This inequality can be rewritten as:
\begin{align}
    y_{K_\Theta k} - \alpha_k \sqrt{\mathrm{Var}[Y_{K_\Theta k}]} \leq E[Y_{K_\Theta k}] \leq y_{K_\Theta k} + \alpha_k \sqrt{\mathrm{Var}[Y_{K_\Theta k}]}
\end{align}
Substituting the expressions for $E[Y_{K_\Theta k}]$ and $\mathrm{Var}[Y_{K_\Theta k}]$ from Lemma~2 turns this inequality into a quadratic inequality in the unknown $\left\| \Delta_{K_\Theta k} \right\|_F^2$. Solving the quadratic gives lower and upper bounds for $\left\| \Delta_{K_\Theta k} \right\|_F^2$, denoted $\underline{\left\| \Delta_{K_\Theta k} \right\|_F^2}$ and $\overline{\left\| \Delta_{K_\Theta k} \right\|_F^2}$. The full derivation is given in Appendix~C of \cite{beheshti2020k}.

\subsection{Bounds on Central Description Length (CDL)}
With these bounds on $\left\| \Delta_{K_\Theta k} \right\|_F^2$, the expected central error $E[Z_{K_\Theta k}]$ can be bounded through Lemma~1. We write $\overline{E[Z_{K_\Theta}]}$ for the resulting upper bound on the expected central error.

A second application of Chebyshev's inequality, now to the central error random variable $Z_{K_\Theta}$, gives
\begin{align}
    P\!\left(|Z_{K_\Theta} - E[Z_{K_\Theta}]| \leq \beta \sqrt{\mathrm{Var}[Z_{K_\Theta}]}\right) \geq 1 - \frac{1}{\beta^2},
\end{align}
so with confidence probability $P_c = 1 - 1/\beta^2$,
\begin{align}
    \overline{z_{K_\Theta}} = E[Z_{K_\Theta}] + \beta \sqrt{\mathrm{Var}[Z_{K_\Theta}]}.
\end{align}
Substituting the upper bound on $E[Z_{K_\Theta}]$ (built from the upper bound on $\left\| \Delta_{K_\Theta k} \right\|_F^2$) and the expression for $\mathrm{Var}[Z_{K_\Theta}]$ from Lemma~1 produces the computable upper bound $\overline{z_{K_\Theta}}$ used as the CDL-CVI loss. In this last step the unknown scatter factor covariances $\overline{\Lambda}$ are replaced by the empirical cluster covariances $\widehat{\Sigma}_{K_\Theta}(k)$.

\end{document}